\documentclass[letterpaper, 10 pt, journal, table]{IEEEtran}

\IEEEoverridecommandlockouts                             
\usepackage{etex}
\usepackage[numbers,compress]{natbib}
\usepackage{multicol}
\usepackage[bookmarks=true,hidelinks]{hyperref}
\usepackage[font={small}]{caption}

\usepackage{comment}
\usepackage{siunitx}
\usepackage{relsize}
\usepackage{ifthen}
\usepackage[colorinlistoftodos]{todonotes}

\usepackage[caption=false,position=bottom]{subfig}
\usepackage[]{caption}

\usepackage{graphics} 
\usepackage{rotating}
\usepackage{color}
\usepackage{enumerate}
\usepackage[T1]{fontenc}
\usepackage{psfrag}
\usepackage{epsfig} 
\usepackage{booktabs}
\usepackage{graphicx,url}
\usepackage{multirow}
\usepackage{array}
\usepackage{latexsym}
\usepackage{amsfonts}
\usepackage{amsmath}
\usepackage{amssymb}
\usepackage{xstring}
\usepackage{mathtools}
\usepackage{multirow}
\usepackage{prettyref}
\usepackage{flexisym}
\usepackage{bigdelim}
\usepackage{breqn} 
\usepackage{listings}
\let\labelindent\relax
\usepackage{enumitem}
\usepackage{xspace}
\usepackage{bm}
\graphicspath{{./figures/}}
\usepackage{tikz}
\usetikzlibrary{matrix,calc}

\usepackage{algorithmicx}
\usepackage{algorithm}
\algrenewcommand\algorithmicindent{0.5em}
\algnewcommand\algorithmicinput{\textbf{Input:}}
\algnewcommand\INPUT{\item[\algorithmicinput]}
\usepackage[]{algpseudocode}

\usepackage{mdwlist}

\let\itemize\undefined
\makecompactlist{itemize}{stditemize}

\usepackage{amsthm}
\newtheoremstyle{mystyle}
  {}
  {}
  {\itshape}
  {}
  {\bfseries}
  {.}
  { }
  {\thmname{#1}\thmnumber{ #2}\thmnote{ (#3)}}
\theoremstyle{mystyle}

\newrefformat{prob}{Problem\,\ref{#1}}
\newrefformat{def}{Definition\,\ref{#1}}
\newrefformat{sec}{Section\,\ref{#1}}
\newrefformat{sub}{Section\,\ref{#1}}
\newrefformat{prop}{Proposition\,\ref{#1}}
\newrefformat{app}{Appendix\,\ref{#1}}
\newrefformat{alg}{Algorithm\,\ref{#1}}
\newrefformat{cor}{Corollary\,\ref{#1}}
\newrefformat{thm}{Theorem\,\ref{#1}}
\newrefformat{lem}{Lemma\,\ref{#1}}
\newrefformat{fig}{Fig.\,\ref{#1}}
\newrefformat{tab}{Table\,\ref{#1}}

\newcommand{\bdmath}{\begin{dmath}}
\newcommand{\edmath}{\end{dmath}}
\newcommand{\beq}{\begin{equation}}
\newcommand{\eeq}{\end{equation}}
\newcommand{\bdm}{\begin{displaymath}}
\newcommand{\edm}{\end{displaymath}}
\newcommand{\bea}{\begin{eqnarray}}
\newcommand{\eea}{\end{eqnarray}}
\newcommand{\beal}{\beq \begin{array}{ll}}
\newcommand{\eeal}{\end{array} \eeq}
\newcommand{\beas}{\begin{eqnarray*}}
\newcommand{\eeas}{\end{eqnarray*}}
\newcommand{\ba}{\begin{array}}
\newcommand{\ea}{\end{array}}
\newcommand{\bit}{\begin{itemize}}
\newcommand{\eit}{\end{itemize}}
\newcommand{\ben}{\begin{enumerate}}
\newcommand{\een}{\end{enumerate}}

\newcommand{\calR}{{\cal R}}

\newcommand{\calX}{{\cal X}}

\newcommand{\etal}{\emph{et~al.}\xspace}
\newcommand{\setal}{~\emph{et~al.}\xspace}
\newcommand{\eg}{\emph{e.g.,}\xspace}
\newcommand{\ie}{\emph{i.e.,}\xspace}
\newcommand{\myParagraph}[1]{{\bf #1.}\xspace}

\newcommand{\M}[1]{{\bm #1}} 
\renewcommand{\boldsymbol}[1]{{\bm #1}}

\newcommand{\hide}[1]{}

\newcommand{\hiddenText}{{\color{gray} hidden text.}}
\newcommand{\hideWithText}[1]{\hiddenText}

\DeclareMathOperator*{\argmin}{arg\,min}

\newcommand{\norm}[1]{\left\| #1 \right\|}

\newcommand{\Real}[1]{ { {\mathbb R}^{#1} } }

\newcommand{\SEthree}{\ensuremath{\mathrm{SE}(3)}\xspace}

\newcommand{\SOthree}{\ensuremath{\mathrm{SO}(3)}\xspace}

\newcommand{\MM}{\M{M}}

\newcommand{\MR}{\M{R}}

\newcommand{\MX}{\M{X}}

\newcommand{\vg}{\boldsymbol{g}}

\newcommand{\vv}{\boldsymbol{v}}
\newcommand{\vt}{\boldsymbol{t}}

\newcommand{\scenario}[1]{{\smaller \sf#1}\xspace}

\newcommand{\blue}[1]{{\color{blue}#1}}

\newcommand{\red}[1]{{\color{red}#1}}

\newcommand{\linkToPdf}[1]{\href{#1}{\blue{(pdf)}}}
\newcommand{\linkToPpt}[1]{\href{#1}{\blue{(ppt)}}}
\newcommand{\linkToCode}[1]{\href{#1}{\blue{(code)}}}
\newcommand{\linkToWeb}[1]{\href{#1}{\blue{(web)}}}
\newcommand{\linkToVideo}[1]{\href{#1}{\blue{(video)}}}
\newcommand{\linkToMedia}[1]{\href{#1}{\blue{(media)}}}
\newcommand{\award}[1]{\xspace} 

\newcommand{\kimera}{\scenario{Kimera}}
\newcommand{\kimeraMulti}{\scenario{Kimera-Multi}} 

\newcommand{\medfield}{\scenario{Medfield}}
\newcommand{\Stata}{\scenario{Stata}}
\newcommand{\camp}{\scenario{Camp}}
\newcommand{\city}{\scenario{City}}
\newcommand{\euroc}{\scenario{EuRoc}}
\newcommand{\RBCD}{RBCD\xspace}

\newcommand{\kimeraVIO}{\scenario{Kimera-VIO}}
\newcommand{\kimeraSemantics}{\scenario{Kimera-Semantics}}

\newcommand{\robot}{\calR}
\newcommand{\DGNC}{D-GNC\xspace} 
\graphicspath{{./figures/}}

\usepackage{float}
\usepackage{hyperref}
\usepackage{graphicx}
\usepackage{epstopdf}
\usepackage{epsfig}

\usepackage{xr}

\definecolor{lime}{HTML}{A6CE39}
\DeclareRobustCommand{\orcidicon}{
	\hspace{-3mm}
	\begin{tikzpicture}
		\draw[lime, fill=lime] (0,0) 
		circle [radius=0.16] 
		node[white] {{\fontfamily{qag}\selectfont \tiny ID}};
		\draw[white, fill=white] (-0.0625,0.095) 
		circle [radius=0.007];
	\end{tikzpicture}
	\hspace{-2mm}
}
\foreach \x in {A, ..., Z}{%
	\expandafter\xdef\csname orcid\x\endcsname{\noexpand\href{https://orcid.org/\csname orcidauthor\x\endcsname}{\noexpand\orcidicon}}
}

\title{\huge{Kimera-Multi: Robust, Distributed, Dense \\ Metric-Semantic SLAM for Multi-Robot Systems}}

\author{
	Yulun Tian\orcidA{}, 
	Yun Chang\orcidB{}, 
	Fernando Herrera Arias\orcidC{}, 
	Carlos Nieto-Granda\orcidD{}, \\ 
	Jonathan P. How\orcidE{}, 
	and Luca Carlone\orcidF{}
\thanks{ 
	Y.\,Tian, Y.\,Chang, F.\,Herrera~Arias, J.P.\,How, and L.\,Carlone are with the Laboratory for Information \& Decision Systems, Massachusetts Institute of Technology, Cambridge, MA, 
  USA,~{\sf \{yulun,yunchang,luisfer,} {\sf jhow,lcarlone\}@mit.edu}.
	C.\,Nieto-Granda is with the U.S. Army Combat Capabilities Development Command, Army Research Laboratory, Adelphi, MD, USA,
	{\sf cnietogr@mit.edu}.
}
\thanks{This work was supported in part by 
ARL Distributed and Collaborative Intelligent Systems and Technology Collaborative Research Alliance (DCIST CRA) under agreement W911NF-17-2-0181, 
in part by ONR under BRC award N000141712072, 
in part by Lincoln Laboratory's Resilient Perception in Degraded Environment Program, 
in part by Carlone's Amazon Research Award, and in part by Mathworks.
}
}

\begin{document}

\maketitle
\begin{tikzpicture}[overlay, remember picture]
	\path (current page.north east) ++(-4.2,-0.2) node[below left] { 
		This paper has been accepted for publication in the IEEE Transactions on Robotics.
	};
\end{tikzpicture}
\begin{tikzpicture}[overlay, remember picture]
	\path (current page.north east) ++(-3.0,-0.6) node[below left] {
		Please cite the paper as: Y.~Tian, Y.~Chang, F.~Herrera Arias, C.~Nieto-Granda, J.~P.~How, and L.~Carlone,
	};
\end{tikzpicture}
\begin{tikzpicture}[overlay, remember picture]
	\path (current page.north east) ++(-3.5,-1) node[below left] {
		``Kimera-Multi: Robust, Distributed, Dense Metric-Semantic SLAM for Multi-Robot Systems'',
	};
\end{tikzpicture}
\begin{tikzpicture}[overlay, remember picture]
	\path (current page.north east) ++(-7.3,-1.4) node[below left] {
		\emph{IEEE Transactions on Robotics (T-RO)}, 2022.
	};
\end{tikzpicture}
\vspace{-4mm}


\begin{abstract}
Multi-robot Simultaneous Localization and Mapping (SLAM) is a crucial capability to obtain timely situational awareness over large areas.
Real-world applications demand multi-robot SLAM systems to be robust to perceptual aliasing and to operate under limited communication bandwidth; moreover, it is desirable for these systems to capture semantic information 
to enable high-level decision-making and spatial AI.

This paper presents \kimeraMulti, a multi-robot system that 
(i) is robust and capable of identifying and rejecting incorrect inter and intra-robot loop closures resulting from perceptual aliasing,
(ii) is fully distributed and only relies on local (peer-to-peer) communication to achieve distributed localization and mapping,
and (iii) builds a globally consistent metric-semantic 3D mesh model of the environment in real-time, where faces of the mesh are annotated with semantic labels.
\kimeraMulti is implemented by a team of robots equipped with visual-inertial sensors. 
Each robot builds a local trajectory estimate and a local mesh using \kimera. 
When communication is available, robots initiate a distributed place recognition and robust pose graph optimization protocol based on a distributed graduated non-convexity algorithm.
The proposed protocol allows the robots to improve their local trajectory estimates by leveraging inter-robot loop closures while being robust to outliers. Finally, each robot uses its improved trajectory estimate to correct the local mesh using mesh deformation techniques. 

We demonstrate \kimeraMulti in photo-realistic simulations, SLAM benchmarking datasets, and challenging outdoor datasets collected using ground robots.
Both real and simulated experiments involve long trajectories (\eg up to 800 meters per robot).
The experiments show that \kimeraMulti (i) outperforms the state of the art in terms of robustness and accuracy, 
(ii) achieves estimation errors comparable to a centralized SLAM system while being fully distributed, 
(iii) is parsimonious in terms of communication bandwidth, 
(iv) produces accurate metric-semantic 3D meshes, and 
(v) is modular and can be also used for standard 3D reconstruction (\ie without 
semantic labels) or for trajectory estimation (\ie  without reconstructing a 3D mesh).
\end{abstract}

\vspace{-4mm}
\section*{Supplementary Material}
Video demonstration of the proposed \kimeraMulti system
on the \medfield dataset (Fig.~\ref{fig:real-world-experiment:medfield}):
\url{https://youtu.be/G7I3JubdU8E}.
\vspace{-1.5mm}

\section{Introduction}
\label{sec:intro}
\vspace{-1.5mm}

\begin{figure}[th]
	\centering
	\subfloat[\kimeraVIO trajectory estimate]{%
		\includegraphics[width=0.46\textwidth]{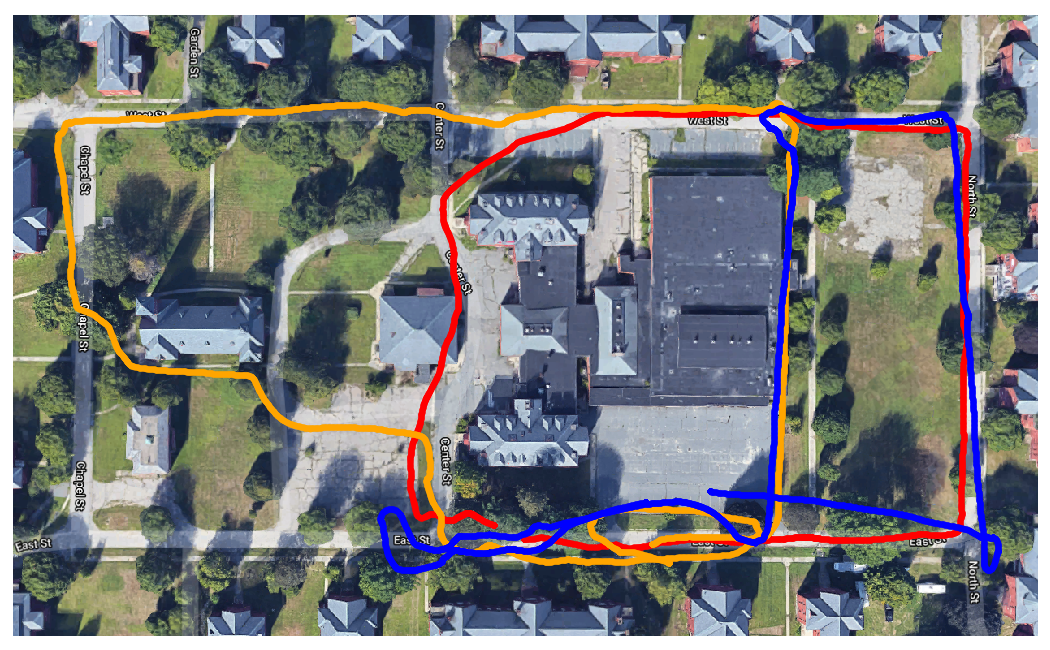}
		\label{fig:medfield_vio_traj}
	}
	\\
	\subfloat[\kimeraMulti trajectory estimate]{%
		\includegraphics[width=0.46\textwidth]{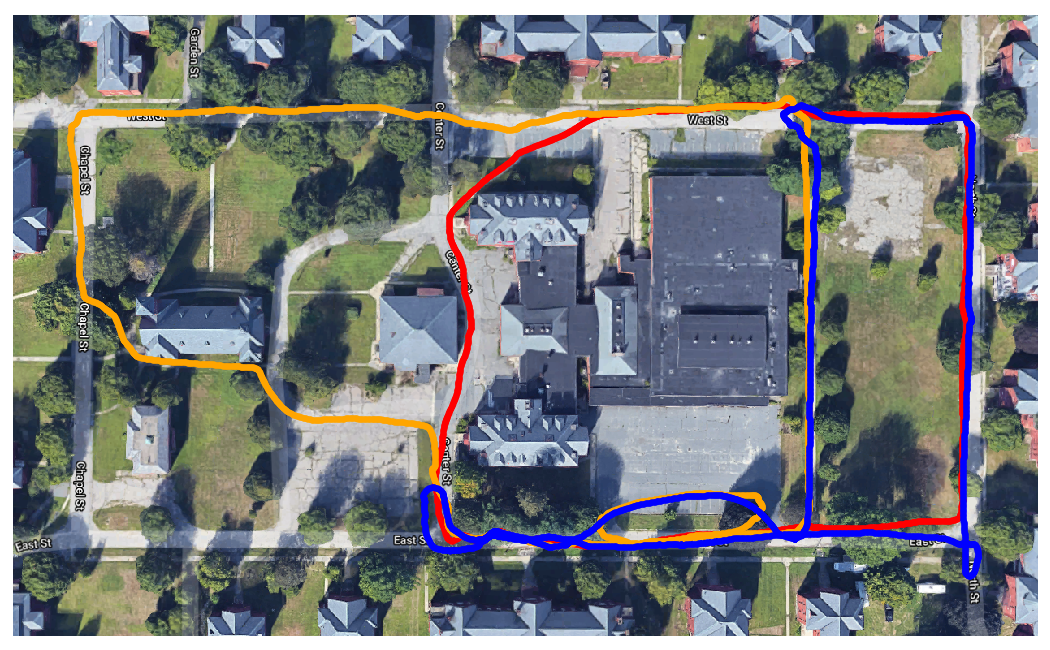}
		\label{fig:medfield_multi_traj}
	}
	\\
	\subfloat[\kimeraMulti optimized mesh]{%
		\includegraphics[trim=90 50 90 70, clip,width=0.46\textwidth]{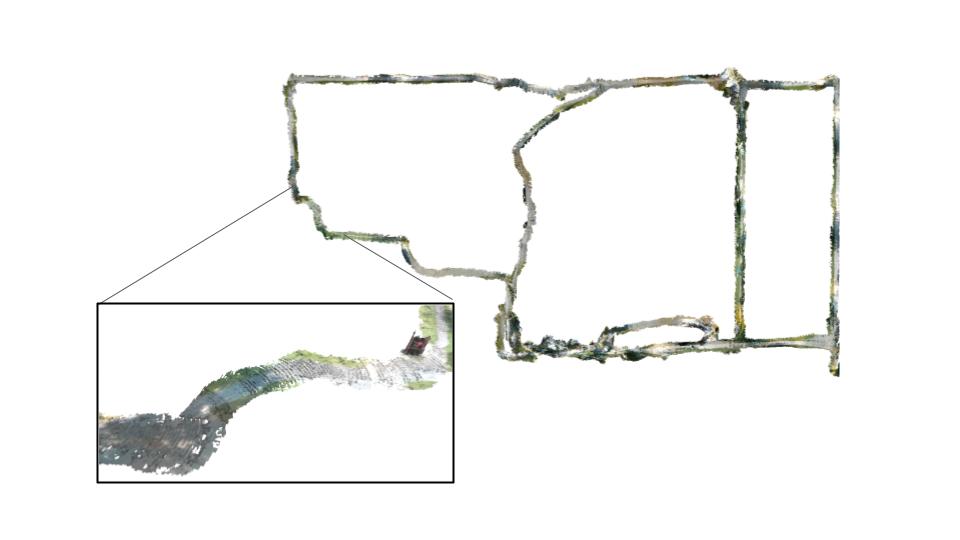}
		\label{fig:medfield_multi_mesh}
	}
	\caption{
		Demonstration of \kimeraMulti in a three-robot collaborative SLAM dataset collected at Medfield, Massachusetts, USA. 
		Total trajectory length (including all robots) is 2188 meters.
		(a) Trajectory estimate from \kimeraVIO is affected by estimation drift.
		(b) \kimeraMulti achieves accurate and robust trajectory estimation.
		(c) \kimeraMulti also produces an optimized 3D mesh of the environment.
		\vspace{-20pt}
		}
	\label{fig:real-world-experiment:medfield}
\end{figure}

Multi-robot collaborative simultaneous localization and mapping (SLAM) is an important topic in robotics research, due to its capability to provide situational awareness over large-scale environments for extended periods of time.
Such capability is fundamental for various applications such as factory automation, search \& rescue, intelligent transportation, planetary exploration, and surveillance and monitoring in military and civilian endeavors.

In this work, we advance state-of-the-art collaborative SLAM by 
developing a system that enables real-time estimation of \emph{dense metric-semantic} 3D mesh models under realistic constraints on communication bandwidth.
The 3D mesh captures the complete and dense geometry of the environment that the robots operate in.
Furthermore, by annotating the mesh with human-understandable semantic labels (\eg ``building'', ``road'', ``object''), our system provides high-level abstractions of the environment that are necessary to enable next-generation \emph{spatial perception}~\citep{Rosinol21ijrr-Kimera} (or Spatial AI~\cite{Davison18arxiv-futureMapping}) and high-level decision making. 
In single-robot SLAM, metric-semantic models have been employed in pioneering work such as SLAM++~\cite{Salas-Moreno13cvpr} and SemanticFusion~\cite{McCormac17icra-semanticFusion}. 
Recent work includes systems that can build metric-semantic 3D models in real-time using a multi-core CPU, including  
Kimera~\cite{Rosinol21ijrr-Kimera} and Voxblox++~\cite{Grinvald19ral-voxbloxpp}.
In multi-robot SLAM, many existing systems rely on sparse landmarks (\eg \citep{Cunningham10iros,Cieslewski18icra}).
While these systems excel at collaborative localization, they do not provide a complete solution to dense mapping, which is required by critical navigation tasks such as collision avoidance and motion planning.
On the other hand, 
recent multi-robot systems begin to leverage semantic information to aid collaborative SLAM, but the underlying representations are still sparse (\eg objects \citep{Choudhary17ijrr-distributedPGO3D,Tchuiev20ral}).
Recent work \citep{Yue20Semantic} employs dense semantic segmentation, but the approach is limited to pairwise matching of local maps.
Overall, there has not been a complete multi-robot system for dense metric-semantic SLAM, partially due to the additional communication and computation costs involved in building such a model.
This work closes this gap by developing a collaborative metric-semantic SLAM system.
Furthermore, the proposed system is fully distributed and is capable of operating under realistic communication constraints.

In addition, this work aims to improve the \emph{robustness} of collaborative SLAM for operations in challenging real-world environments. 
In practice, perceptual aliasing caused by similar-looking scenes often results in wrong inter-robot data associations (\ie outlier loop closures), which in turn cause catastrophic failures of standard estimation back-ends.
In multi-robot SLAM, this issue is further complicated by the lack of a common reference frame and a global outlier-free odometry backbone.
While recent work has proposed several robust estimation techniques for collaborative SLAM, they either rely too heavily on initialization \cite{Indelman14icra,Dong15icra} or employ heuristic search methods \cite{Mangelson18icra,Lajoie20ral-doorSLAM} which can cause low recall (\ie missing correct loop closures).
This work addresses this challenge by developing a robust distributed back-end based on graduated non-convexity (GNC) \citep{Yang20ral-GNC}. 

\myParagraph{Contributions}
The primary contribution of this work is \kimeraMulti, \emph{a fully distributed system for multi-robot dense metric-semantic SLAM}. 
Our system enables a team of robots to collaboratively estimate a semantically annotated 3D mesh model of the environment in real-time.
Each robot runs \kimera \cite{Rosinol20icra-Kimera} to process onboard visual-inertial sensor data and obtain local trajectory and 3D mesh estimates. When communication becomes available, a fully distributed procedure is triggered to perform inter-robot place recognition, relative pose estimation, and robust distributed trajectory estimation.
From the jointly optimized trajectory estimates, each robot performs real-time local mesh deformation to correct local mapping drift and improve global map consistency. 
The implementation of \kimeraMulti is modular and allows different components to be disabled or replaced.
Fig.~\ref{fig:real-world-experiment:medfield} demonstrates \kimeraMulti on a three-robot collaborative SLAM dataset collected at Medfield, Massachusetts, USA.

The second technical contribution of this work is a \emph{new, two-stage method for outlier-robust distributed pose graph optimization (PGO)}, which serves as the distributed back-end of \kimeraMulti. 
The first stage initializes robots' local trajectories in a global reference frame by using GNC \cite{Yang20ral-GNC} to estimate relative transformations between the coordinate frames of pairs of robots. 
This method is robust to outlier loop closures and, furthermore, is efficient because it does not require iterative communication.
The second stage solves the full robust PGO problem.
For this purpose, we present a distributed extension of GNC built on top of the state-of-the-art \RBCD solver \cite{tian2019distributed}.
Compared to prior techniques, our approach achieves more robust and accurate trajectory estimation, and is less sensitive to parameter tuning. 

Our third contribution is \emph{an extensive experimental evaluation}. 
We present quantitative evaluations of \kimeraMulti on a collection of large-scale photo-realistic simulations and SLAM benchmarking datasets.
In addition, we demonstrate \kimeraMulti on challenging real-world datasets collected by autonomous ground robots.
Our results show that \kimeraMulti 
(i) provides more robust and accurate distributed trajectory estimation compared to alternative techniques employed in prior work \cite{Mangelson18icra,Lajoie20ral-doorSLAM},
(ii) achieves estimation accuracy that is similar to a centralized system while being fully distributed,
(iii) is communication-efficient and achieves as much as 70\% communication reduction compared to baseline centralized systems,
(iv) builds accurate metric-semantic 3D meshes, 
(v) is modular and allows different features such as mesh reconstruction and semantic annotation to be disabled according to user needs.

\myParagraph{Novelty with respect to previous work \cite{chang2020kimeramulti}}
An earlier version of \kimeraMulti was presented in \cite{chang2020kimeramulti}, but the
present paper extends that work with two new contributions.
First, we develop an \emph{outlier-robust} and \emph{fully distributed} trajectory estimation method based on GNC. 
In \citep{chang2020kimeramulti}, we used an incremental extension of Pairwise Consistency Maximization (PCM)~\citep{Mangelson18icra} to reject outlier loop closures.
However, PCM employs a graph-theoretic formulation and in practice relies on heuristic maximum clique search which causes low recall.
In this work, we show that the proposed distributed GNC method outperforms PCM in terms of robustness and accuracy, and is less sensitive to parameter tuning. 
The second new contribution is a set of additional experimental evaluations.
These include a comprehensive evaluation of different robust distributed PGO techniques (Section~\ref{sec:robustness_experiment}),
evaluations on additional photo-realistic simulations and benchmarking datasets (Section~\ref{sec:dataset_evaluation}), 
and evaluations on two new challenging outdoor datasets (Fig.~\ref{fig:real-world-experiment:medfield} and Fig.~\ref{fig:real-world-experiment:stata}) collected using autonomous ground robots (Section~\ref{sec:outdoor_datasets}).

\section{Related Work}
\label{sec:relatedWork}

\subsection{Metric-Semantic SLAM}
In recent years, single-robot SLAM research is steadily moving towards systems that can build 
\emph{metric-semantic} maps~\cite{Tateno17cvpr-CNN-SLAM,Lianos18eccv-VSO,Dong17cvpr-XVIO,Behley19iccv-semanticKitti,McCormac17icra-semanticFusion,Zheng19arxiv-metricSemantic,Tateno15iros-metricSemantic,Li16iros-metricSemantic,McCormac183dv-fusion++,Runz18ismar-maskfusion,Runz17icra-cofusion,Xu19icra-midFusion,Rosinol20icra-Kimera,Grinvald19ral-voxbloxpp,Rosinol20rss-dynamicSceneGraphs}.
Related research efforts include systems building voxel-based models~\cite{McCormac17icra-semanticFusion,Zheng19arxiv-metricSemantic,Tateno15iros-metricSemantic,Li16iros-metricSemantic,McCormac183dv-fusion++,Runz18ismar-maskfusion,Runz17icra-cofusion,Xu19icra-midFusion}, 
ESDF and meshes~\cite{Rosinol20icra-Kimera,Rosinol19icra-incremental,Grinvald19ral-voxbloxpp}, or 3D scene graphs~\cite{Rosinol20rss-dynamicSceneGraphs}. 
In this work, we build our multi-robot metric-semantic SLAM system on top of 
\kimera~\cite{Rosinol20icra-Kimera}, which provides accurate real-time visual-inertial odometry (VIO) and lightweight mesh reconstruction.

In the multi-robot SLAM literature, the majority of approaches
have focused on dense geometric representations 
(\eg occupancy maps~\cite{Schuster19JFR})
or sparse landmark maps~\citep{Cunningham10iros,Cunningham13icra}; 
see \cite{SajadSaeedi2016MultipleRobotSL} and the references therein. 
Recent work begins to incorporate sparse objects or dense semantic
information in multi-robot perception.
Choudhary~\etal~\citep{Choudhary17ijrr-distributedPGO3D} use class labels to associate objects within a multi-robot pose graph SLAM framework.
Tchuiev and Indelman~\citep{Tchuiev20ral} develop a distributed object-based SLAM method that leverages the coupling between object classification and pose estimation.
Yue~\etal~\citep{Yue20Semantic} leverage dense semantic segmentation 
to perform relative localization and map matching between pairs of robots.
In \citep{chang2020kimeramulti}, we presented an early version of \kimeraMulti and demonstrated it as the first fully distributed system for multi-robot dense metric-semantic SLAM.
The present paper extends \citep{chang2020kimeramulti} with a new \emph{outlier-robust} distributed pose graph optimization algorithm and additional experimental evaluations.

\subsection{Distributed Loop Closure Detection}
Inter-robot loop closures are critical to align the trajectories of the robots in a common reference frame and to improve their trajectory estimates.
In a centralized visual SLAM system (\eg \citep{Schmuck18CCM}), 
robots transmit a combination of global descriptors
(\eg bag-of-words vectors~\citep{Sivic03iccv,Galvez12tro-dbow} and learned full-image descriptors~\citep{Arandjelovic16cvpr-netvlad})
and local visual features (\eg \cite{Lowe99iccv,Bay06eccv}) to a central server that performs centralized place recognition and geometric verification. 
Recent work develops \emph{distributed} and \emph{communication-efficient} paradigms for inter-robot loop closure detection.
Cieslewski and Scaramuzza \cite{Cieslewski17Inverted} propose an efficient method for distributed visual place recognition, based on splitting and distributing bag-of-words visual features~\cite{Sivic03iccv}.
A subsequent approach is developed in \citep{Cieslewski17Netvlad, Cieslewski18icra} based on clustering and distributing NetVLAD~\cite{Arandjelovic16cvpr-netvlad} descriptors. 
A complementary line of work develops efficient methods for distributed geometric verification. 
Giamou~\etal~\cite{giamou2018talk} develop a method to verify a set of candidate inter-robot loop closures using minimum data exchange.
Tian~\etal~\cite{tian2018near,tian2019resource} consider distributed geometric verification under communication and computation budgets and develop near-optimal communication policies based on submodular optimization.   

\subsection{Distributed PGO}
Pose graph optimization (PGO) is commonly used as the estimation backbone of state-of-the-art SLAM systems.
Centralized approaches for multi-robot PGO collect all measurements
at a central station, which computes the trajectory estimates for all the
robots~\cite{Andersson08icra,Kim10icra,Bailey11icra,Lazaro11icra,Dong15icra}.
In parallel, considerable efforts have been made to design \emph{distributed} PGO methods.
Cunningham~\etal~\cite{Cunningham10iros,Cunningham13icra} use Gaussian elimination to exchange marginals over the separator poses.
Another family of approaches are based on distributed gradient descent~\cite{Knuth13icra,Tron2014CameraNetwork,cristofalo2020geod}.
Aragues\setal~\cite{Aragues11icra-distributedLocalization} use a distributed Jacobi approach to estimate 2D poses. 
Choudhary~\etal~\citep{Choudhary17ijrr-distributedPGO3D} propose a two-stage approach that uses distributed Gauss-Seidel method to initialize rotation estimates and solve a single Gauss-Newton iteration.
This method is also implemented as the distributed back-end in recent decentralized SLAM systems~\citep{Cieslewski18icra,Lajoie20ral-doorSLAM}. 
Recently, Fan and Murphey~\citep{fan2020majorization} propose a majorization-minimization method that adapts Nesterov's
acceleration technique to achieve significant empirical speedup. 
Tian~\etal~\citep{tian2019distributed} develop the Riemannian block-coordinate descent (RBCD) method that can be similarly accelerated, and furthermore propose a distributed global optimality verification method based on accelerated power iteration. 
The conference version of \kimeraMulti~\citep{chang2020kimeramulti} uses RBCD as the distributed back-end.
A subsequent work~\citep{tian2020asynchronous} develops distributed PGO with convergence guarantees under asynchronous communication.

\subsection{Robust PGO}
Standard least squares formulation of PGO is susceptible to outlier loop closures that can severely impact trajectory estimation.
To mitigate the effect of outliers in single-robot SLAM, 
early methods are based on RANSAC~\cite{Fischler81}, branch \&
bound~\cite{Neira01tra}, and M-estimation (\cite{Bosse17fnt,Hartley13ijcv}).
{S\"underhauf and Protzel~\cite{Sunderhauf12iros} develop a method to deactivate outliers using binary variables.}
{Agarwal~\etal~\cite{Agarwal13icra} build on the same idea and develop the dynamic covariance scaling method.} 
{Hartley~\etal~\citep{Hartley11cvpr-l1rotationaveraging} and Casafranca~\etal~\citep{Casafranca13iros} 
propose to minimize the $\ell1$-norm of residual errors.
}
{Chatterjee and Govindu~\citep{Chatterjee13iccv,Chatterjee18pami-rotationAveraging}
develop iteratively reweighted least squares (IRLS) methods to solve rotation averaging using a family of robust cost functions.}
{Hu~\etal~\citep{Hu13Reliable} develop similar IRLS methods for single-robot SLAM.}
{Olson and Agarwal~\cite{Olson12rss} and Pfingsthorn and
Birk~\cite{Pfingsthorn13ijrr,Pfingsthorn16ijrr} consider multi-modal
distributions for the noise. }
{Lajoie~\etal~\cite{Lajoie19ral-DCGM} and Carlone and
Calafiore~\cite{Carlone18ral-robustPGO2D} develop {global} solvers based on convex relaxations.}
{A separate line of work investigates {consensus maximization} formulations
that seek to identify the maximal set of mutually consistent inliers~\cite{Carlone14iros-robustPGO2D,Graham15iros,Antonante21tro-outlierRobustEstimation}.}
{Recently, Yang~\etal~\cite{Yang20ral-GNC} develop graduated non-convexity (GNC) that optimizes a sequence of increasingly non-convex surrogate cost functions, and demonstrate state-of-the-art performance on robust PGO problems.}

{In multi-robot SLAM, 
Indelman~\etal~\cite{Indelman14icra} and Dong~\etal~\cite{Dong15icra} apply expectation-maximization to find consistent inter-robot loop closures and estimate initial relative transformations between robots.
{Mangelson~\etal~\cite{Mangelson18icra} design the Pairwise Consistency Maximization (PCM) approach to perform robust map merging between pairs of robots.}
{Lajoie~\etal~\cite{Lajoie20ral-doorSLAM} develop the DOOR-SLAM system which implements an extended version of PCM as the outlier rejection method before distributed trajectory estimation.}
{The recent NeBula system~\citep{Agha21Nebula} also employs PCM within a centralized collaborative SLAM architecture.}
{In this work, we develop a fully distributed extension of GNC~\citep{Yang20ral-GNC}, and demonstrate that our method outperforms PCM in terms of robustness and accuracy.}


\section{System Overview}

\begin{figure}[t]
	\centering
	\includegraphics[width=0.99\columnwidth,
	trim=10mm 45mm 35mm 25mm,clip]{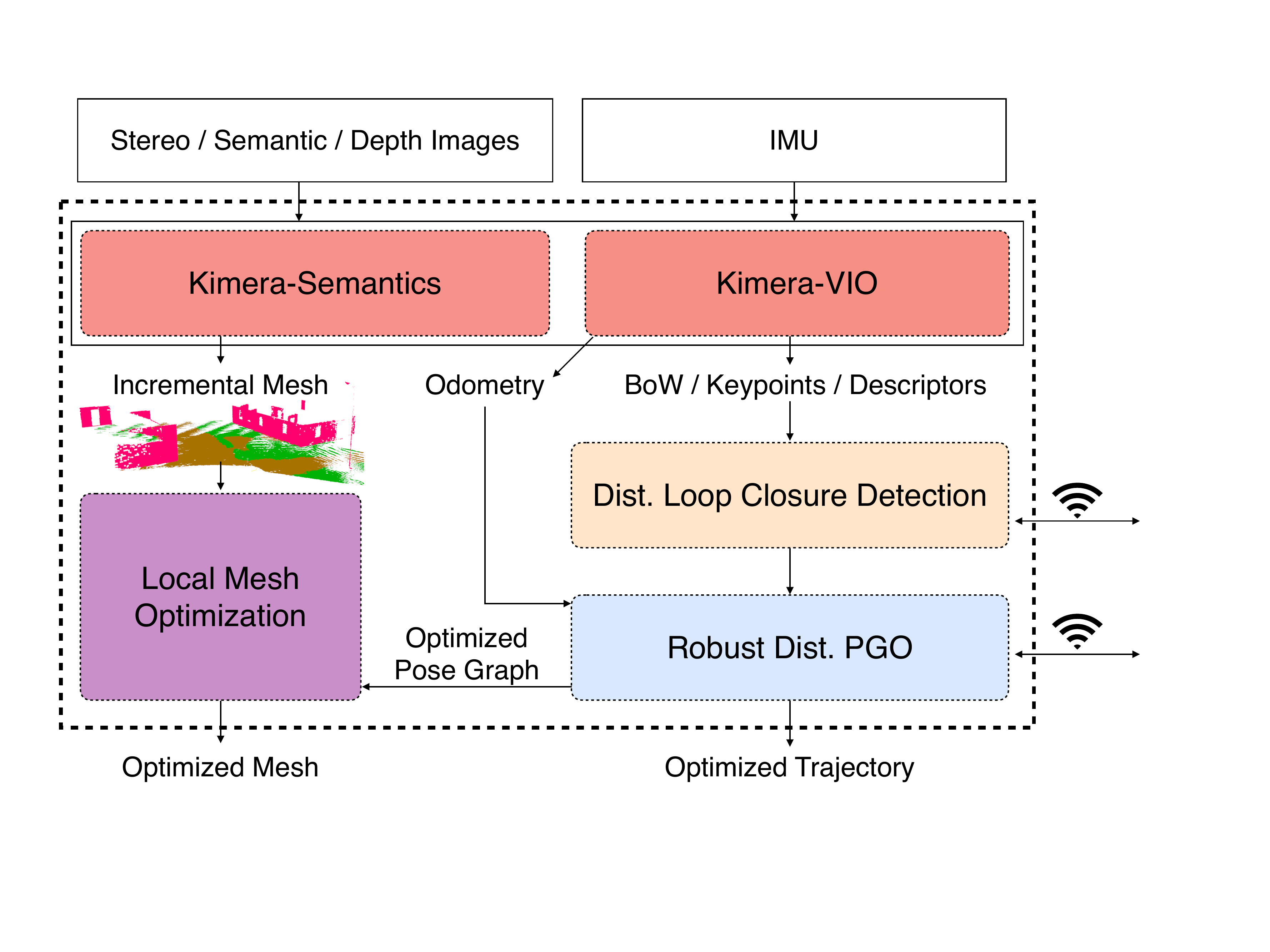}
	\caption{
		\kimeraMulti: system architecture. 
		Each robot runs \kimera (including \kimeraVIO and \kimeraSemantics)
		to estimate local trajectory and mesh. 
		Robots then communicate to perform distributed loop closure detection
		and robust distributed PGO.
		Given the optimized trajectory, each robot performs local mesh optimization.
		\label{fig:architecture} \vspace{-4mm}}
\end{figure}

In \kimeraMulti, each robot runs the \emph{fully decentralized} metric-semantic SLAM system shown in Fig.~\ref{fig:architecture}.
The system consists of four main modules:
(i)~local (single-robot) \kimera, 
(ii)~distributed loop closure detection, 
(iii)~robust distributed trajectory estimation via PGO, 
and (iv) local mesh optimization. 
Among these modules, distributed loop closure detection and robust distributed PGO are the only ones that involve communication between robots.
Fig.~\ref{fig:two_robot_dataflow} shows the data flow between these modules.

{\bf Kimera}~\cite{Rosinol21ijrr-Kimera} runs onboard each robot and provides real-time local trajectory and mesh estimation.
In particular, \kimeraVIO~\cite{Rosinol20icra-Kimera} serves as the visual-inertial odometry module, 
which processes raw stereo images and IMU data to obtain an estimate of the odometric trajectory of the robot. 
\kimeraSemantics~\cite{Rosinol20icra-Kimera} processes depth images (possibly obtained from RGB-D cameras or by stereo matching) 
and 2D semantic segmentations~\cite{GarciaGarcia17arxiv} and produces a dense metric-semantic 3D mesh using the VIO pose estimates. 
In addition, \kimeraVIO computes a Bag-of-Words (BoW) representation of each keyframe using ORB features and DBoW2~\cite{Galvez12tro-dbow}, 
which is used for distributed loop closure detection. 
Interested readers are referred to \cite{Rosinol20icra-Kimera,Rosinol21ijrr-Kimera} for more technical details. 

{{\bf Distributed Loop Closure Detection (Section~\ref{sec:frontend})}} is executed whenever two robots $\alpha$ and $\beta$ are within communication range.
The robots exchange BoW descriptors of the keyframes they collected.
When the robots find a pair of matching descriptors 
(typically corresponding to observations of the same place), 
they perform relative pose estimation using 
standard geometric verification techniques.
The relative pose corresponds to a putative inter-robot loop closure, and is used during robust distributed trajectory estimation. 

{\bf Robust Distributed Trajectory Estimation (Section~\ref{sec:traj_estimation})}
solves for the optimal trajectory estimates of all robots in a global reference frame,
by performing robust distributed PGO using 
odometric measurements from \kimeraVIO
and all putative loop closures detected so far.
At the beginning, a robust initialization scheme is used to find coarse relative transformations between robots' reference frames.
Then, a robust optimization procedure based on a distributed extension of GNC~\citep{Yang20ral-GNC} using the \RBCD solver~\citep{tian2019distributed} is employed to simultaneously select inlier loop closures and recover optimal trajectory estimates. 
Compared to the incremental PCM technique~\citep{Mangelson18icra} used in the conference version of \kimeraMulti~\citep{chang2020kimeramulti},
our new approach enables more robust and accurate trajectory estimation, and is less sensitive to parameter tuning.

{\bf Local Mesh Optimization (Section~\ref{sec:pgmo})} is executed after the robust distributed trajectory estimation stage. 
This module performs a local processing step that deforms the mesh at each robot to enforce consistency with the trajectory estimate resulting from distributed PGO.

\kimeraMulti is 
implemented in C++ and uses the Robot Operating System~(ROS)~\cite{Quigley09icra-ros} 
as a communication layer between robots and between the modules executed on each robot.
The system runs online using a CPU and is modular, thus allowing modules to be replaced or removed.
For instance, the system can also produce a dense \emph{metric} mesh if semantic labels are not available, or only 
produce the optimized trajectory if the dense reconstruction is not required by the user.

\section{Distributed Loop Closure Detection}
\label{sec:frontend}

\begin{figure}[t]
	\centering
	\includegraphics[width=0.90\columnwidth,
	trim=0mm 0mm 0mm 0mm,clip]{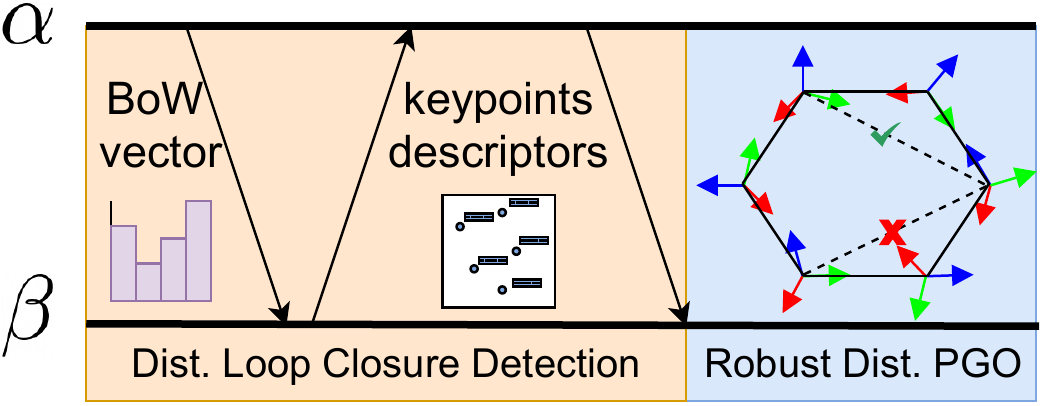}
	\caption{Communication protocol and data flow between pair of robots. \label{fig:two_robot_dataflow} }
	\vspace{-6mm}
\end{figure}

This section describes the front-end of \kimeraMulti, which is responsible for detecting inter-robot loop closures between pairs of robots.
The information flow is summarized in Fig.~\ref{fig:two_robot_dataflow}.
When communication becomes available, robot $\alpha$ initiates the distributed loop closure detection process, by sending global descriptors of its new keyframes since the last rendezvous
to the other robot $\beta$.
We implement these descriptors as bag-of-word vectors using the DBoW2 library \cite{Galvez12tro-dbow}. 
Upon receiving the bag-of-word vectors, robot $\beta$ searches within its own keyframes for candidate matches whose normalized visual similarity scores exceed a threshold ($\geq 0.1$ in our code).
When a potential loop closure is identified, the robots perform standard geometric verification to estimate the relative transformation between the two matched keyframes. 
In our implementation, robot $\beta$ first requests the 3D keypoints and associated descriptors of the matched keyframe from robot $\alpha$ (Fig.~\ref{fig:two_robot_dataflow}). 
Subsequently, robot $\beta$ computes putative correspondences by matching the two sets of feature descriptors using nearest neighbor search implemented in 
OpenCV \cite{OpenCV}.
From the putative correspondences, robot $\beta$ attempts to compute the relative transformation using Nist\'er's five-point method~\cite{Nister04pami} and Arun's three-point method~\cite{Arun87pami} combined with RANSAC~\cite{Fischler81}.
Both techniques are implemented in the OpenGV library~\cite{OpenGV}.
If geometric verification succeeds with more than {five correspondences},
the loop closure is accepted and sent to the robust distributed trajectory estimation module. 

\section{Robust Distributed Trajectory Estimation}
\label{sec:traj_estimation}

In \kimeraMulti, the robots estimate their trajectories by collaboratively solving a
PGO problem using the entire team's odometry measurements and intra-robot and inter-robot
loop closures. Some of these loop closures may
be outliers (due to, \eg perceptual aliasing) and thus we need an outlier-robust method for solving PGO.
In the earlier version of \kimeraMulti \cite{chang2020kimeramulti}, we used an 
incremental variant of PCM \cite{Mangelson18icra} for
outlier rejection via maximum clique computation prior to trajectory estimation. 
However, even with parallelization~\citep{Rossi15parallel}, the runtime of exact maximum clique search exceeds 10 seconds already in graphs with 700 loop closures, which is not practical for our application.
For this reason, in practice PCM has to rely on heuristic maximum clique algorithms and thus often exhibits poor recall, as shown in Section~\ref{sec:robustness_experiment}.

In this paper, we propose a new distributed approach for robust trajectory estimation based on GNC \citep{Yang20ral-GNC}.
The main idea in GNC is to start from a convex approximation of the robust cost function and then gradually introduce the non-convexity to prevent convergence to spurious solutions. 
While in general GNC does not require an initial guess~\citep{Yang20ral-GNC}, it has been observed that global solvers for 3D SLAM (\eg SE-Sync~\citep{Rosen18ijrr-sesync}) become too slow in the presence of outliers~\citep{Antonante21tro-outlierRobustEstimation}. 
For this reason, in \citep{Antonante21tro-outlierRobustEstimation} local optimization is performed instead at each iteration of GNC (starting from an outlier-free initial guess), and this approach has been shown to be very effective.
In single-robot SLAM, one can easily obtain an outlier-free initial guess by chaining together odometry measurements.
In the multi-robot case, there is no odometry between different robots' poses, and the challenge thus becomes building an initial guess that is insensitive to outliers.

To address the aforementioned challenge, the proposed distributed graduated non-convexity (\DGNC) approach involves two stages.
In the first stage (Section~\ref{sec:init}), we use an outlier-robust and communication-efficient method to initialize robots' trajectories in a global reference frame.
In the second stage (Section~\ref{sec:dpgo}), we develop a fully distributed procedure to execute GNC, using the \RBCD distributed solver as a subroutine.
Algorithm~\ref{alg:dgnc} provides the pseudocode of \DGNC.

\begin{algorithm}[t]
	\caption{{\small Distributed Graduated Non-Convexity (\DGNC) }}
	\label{alg:dgnc}
	\begin{algorithmic}[1]
		\renewcommand{\algorithmicrequire}{\textbf{Input:}}
		\renewcommand{\algorithmicensure}{\textbf{Output:}}
		\Require 
		\Statex - Initial trajectory estimates in \emph{local} frames of each robot
		\Statex - Odometry and intra-robot and inter-robot loop closures that each robot is involved in
		\Statex - Threshold $\bar{c}$ of truncated least squares (TLS) cost
		\Ensure
		\Statex - Optimized trajectory estimate of each robot in global frame
		\vspace{2pt}
		\State \textbf{Robust initialization}:
		robots communicate to initialize trajectory estimates in a global reference frame (Section~\ref{sec:init}). 
		\label{alg:dgnc:init}
		\State In parallel, each robot initializes GNC weights for its \emph{local} intra and inter-robot loop closures $w_i=1, \forall i$.
		\While{not converged}
			\State \textbf{Variable update}:
			with fixed weights,
			robots communicate to execute \RBCD for $T$ iterations (default $T=15$)\label{alg:dgnc:var_update}.
			\State \textbf{Weight update}:
			in parallel, each robot updates GNC weights for intra-robot loop closures and inter-robot loop closures it is involved in.
			\label{alg:dgnc:weight_update}
			\State \textbf{Parameter update}:
			in parallel, each robot updates the control parameter $\mu$.
			\label{alg:dgnc:mu_update}
		\EndWhile
	\end{algorithmic}
\end{algorithm}

\subsection{Background: Graduated Non-Convexity}
\label{sec:gnc}
We start by providing a brief review of GNC \cite{Yang20ral-GNC,Black96ijcv-unification}. 
One challenge associated with classical M-estimation \cite{Huber81,Zhang97icv} is that the employed robust cost function $\rho$ can be highly non-convex, hence making local search techniques sensitive to the initial guess.
The key idea behind GNC is to optimize a sequence of easier (\ie less non-convex) surrogate cost functions that gradually converges to the original robust cost function.
Each surrogate problem takes the same form as classical M-estimation,
\begin{align}
\min_{x \in \calX} \,\, \sum_{i} \rho_\mu(r_i(x)),
\label{eq:gnc}
\end{align}
where $r_i : \calX \to \mathbb{R}$ is the residual error associated with the $i$th measurement. 
The sequence of surrogate functions $\rho_\mu$, parameterized by control parameter $\mu$, satisfies that for some given constants $\mu_0$ and $\mu_1$:
(i) for $\mu \to \mu_0$, the function $\rho_\mu$ is convex, 
and (ii) for $\mu \to \mu_1$, $\rho_\mu$ converges to the  original (non-convex) robust cost function $\rho$.
In practice, one initializes $\mu$ near $\mu_0$, and gradually updates its value to approach $\mu_1$ as optimization proceeds.

For each instance of \eqref{eq:gnc}, GNC reformulates the problem using 
the Black-Rangarajan Duality~\citep{Black96ijcv-unification},
which states that under certain technical conditions (satisfied by all common choices of robust cost functions), \eqref{eq:gnc} is equivalent to the following optimization problem,
\begin{align}
\min_{x \in \calX, w_i \in [0,1]} \,\, 
\sum_{i} \big [w_i r_i^2(x) + \Phi_{\rho_\mu}(w_i) \big],
\label{eq:outlier_process}
\end{align}
where $w_i \in [0,1]$ is a scalar weight associated with the $i$th measurement.
In \eqref{eq:outlier_process}, the \emph{outlier process} $\Phi_{\rho_\mu}(w_i)$ introduces a penalty term for each $w_i$, and its expression depends on the chosen robust cost function $\rho$ and the control parameter $\mu$. 
Similar to the classical iterative reweighted least squares (IRLS) scheme, 
GNC performs alternating minimization over the variable $x$ and weights $w_i$ to optimize \eqref{eq:outlier_process}, but in the meantime also updates the control parameter $\mu$:
\begin{enumerate}
		\item \textbf{Variable update:} 
		Minimize \eqref{eq:outlier_process} with respect to $x$ with fixed weights $w_i$.
		This amounts to solving a standard weighted least-squares problem,
				\begin{align}
						x^\star \in \argmin_{x \in \calX} \sum_{i} w_i r_i^2(x).
						\label{eq:var_update}
				\end{align}
		\item \textbf{Weight update:} 
		Minimize \eqref{eq:outlier_process} with respect to $w_i$ with fixed variable $x$.
		The corresponding update for each $w_i$ has a closed-form expression that depends on the current robust surrogate function $\rho_\mu$; see \cite[Proposition~3-4]{Yang20ral-GNC}. 
		\item \textbf{Parameter update:}
		Update $\mu$ by a constant factor to approach $\mu_1$.
\end{enumerate}

The control parameter $\mu$ is initialized at a value close to $\mu_0$.
In the absence of a better guess, all weights are initialized to one (\ie all measurements are considered inliers initially).
Then the steps above are repeated until $\mu$ approaches $\mu_1$.

\subsection{Robust Distributed Initialization}
\label{sec:init}

\begin{figure}[t]
	\centering
	\includegraphics[width=0.75\columnwidth,
	trim=0mm 0mm 0mm 0mm,clip]{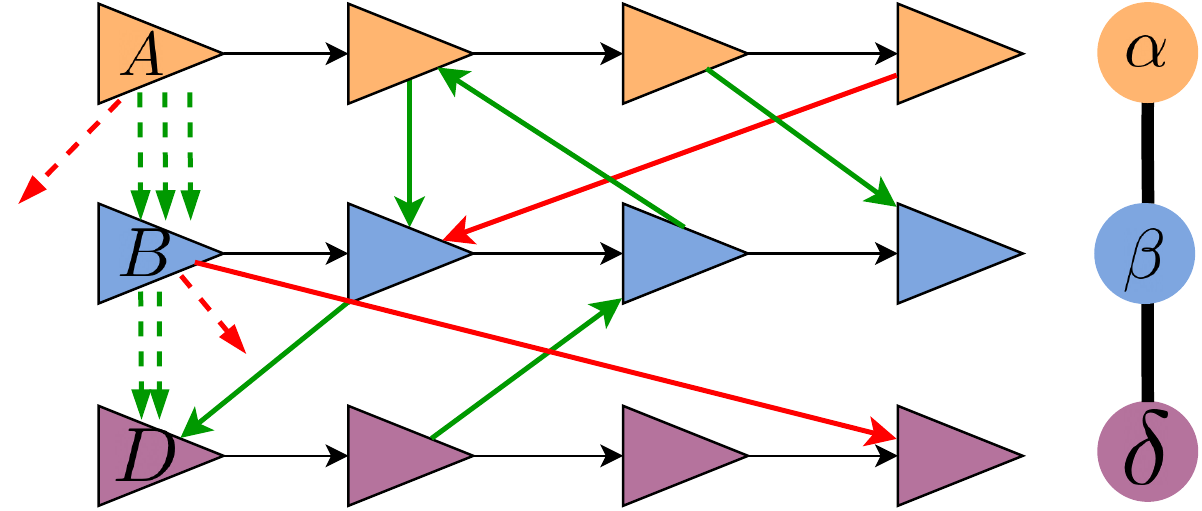}
	\caption{
		Robust distributed initialization.
		\textbf{Left}:
		Three-robot scenario with local reference frames $A,B,D$, each coinciding with the first pose of the corresponding robot.
		Between every pair of robots, inlier loop closures ({\color[HTML]{009900} $\longrightarrow$})
		lead to similar estimates for the alignment between frames ({\color[HTML]{009900} $\dashrightarrow$}). 
		Each outlier loop closure (\red{$\longrightarrow$}) produces an outlier frame alignment (\red{$\dashrightarrow$}), which can be rejected with GNC.
		\textbf{Right}: 
		Corresponding robot-level spanning tree.
		\label{fig:robust_init} }
	\vspace{-6mm}
\end{figure}

To optimize the pose graph, we first need to initialize all robot poses in a shared (global) coordinate frame (Algorithm~\ref{alg:dgnc}, line~\ref{alg:dgnc:init}).
Each robot can readily initialize its trajectory in its
\emph{local} reference frame by chaining odometry measurements. 
To express these local initial guesses in the global reference frame, however, we must estimate the
relative pose between the local reference frames. 

\myParagraph{Pairwise coordinate frame estimation}
First, let us see how this can be done
between two robots $\alpha$ and $\beta$, with local reference frames $A$ and $B$, respectively.
Consider a loop closure between the $i$th pose of $\alpha$ and $j$th pose of $\beta$, denoted as $\widetilde{\MX}^{\alpha_i}_{\beta_j} \in \SEthree$.
Denote the odometric estimates of pose $i$ and $j$ (in the \emph{local} frames of the two robots) as $\widehat{\MX}^A_{\alpha_i},\widehat{\MX}^B_{\beta_j} \in \SEthree$. 
By combining these pose estimates with the loop closure, we obtain a noisy estimate of the relative transformation between frames $A$ and $B$,
\begin{equation}
\widehat{\MX}^{A}_{B_{ij}} \triangleq
\widehat{\MX}^A_{\alpha_i}
\widetilde{\MX}^{\alpha_i}_{\beta_j}
\big(\widehat{\MX}^B_{\beta_j} \big)^{-1},
\label{eq:naive_inter_frame}
\end{equation}
where the subscript of $\widehat{\MX}^{A}_{B_{ij}}$ indicates that this estimate is computed using loop closure $(i,j)$.
From \eqref{eq:naive_inter_frame}, we see that each inter-robot loop closure provides a candidate alignment for the reference frames $A$ and $B$.
Furthermore, candidate alignments produced by inlier loop closures are expected to be in mutual agreement; see Fig.~\ref{fig:robust_init} and also \cite{Indelman14icra}.
To obtain a reliable estimate of the true relative transformation, 
we thus formulate and solve the following \emph{robust} pose averaging problem,
\begin{align}
\widehat{\MX}^A_B & \in \argmin_{\MX \in \SEthree}
\;\; \sum_{(i,j) \in L_{\alpha,\beta}} \rho(r_{ij}(\MX)),
\label{eq:pose_avg}
\end{align}
where $\rho : \mathbb{R} \to \mathbb{R}$ is the truncated least squares (TLS) robust cost function \cite{Yang20ral-GNC}, and $L_{\alpha,\beta}$ is the set of inter-robot loop closures between robot $\alpha$ and $\beta$.
Each residual measures the geodesic distance between the to-be-computed average pose $\MX$ and the measurement $\widehat{\MX}^{A}_{B_{ij}}$,
\begin{align}
r_{ij}(\MX) & \triangleq \left\| 
\MX \boxminus \widehat{\MX}^{A}_{B_{ij}}
\right\|_{\Sigma},
\end{align}
where $\Sigma \in \mathbb{S}_{++}^{6}$ is a fixed covariance matrix.
In our implementation, we use a diagonal covariance with a standard deviation of $0.1$~rad for rotation and $0.5$~m for translation.
Between a given pair of robots, one robot can solve \eqref{eq:pose_avg} \emph{locally} using GNC~\cite{Yang20ral-GNC} without extra communication (since each robot already has access to all loop closures it is involved in), and transmits the solution to the other robot. 
In practice, we use the GNC implementation available in GTSAM~\citep{gtsam}, 
which uses Levenberg-Marquardt (initialized at identity pose) in each GNC variable update to solve \eqref{eq:pose_avg}.

\myParagraph{Multi-robot coordinate frame estimation}
The above \emph{pairwise} procedure can be executed repeatedly to express \emph{all} local reference frames (and trajectory estimates) in a global frame while being robust to outliers.
To do so, we first choose an arbitrary spanning tree in
the robot-level dependency graph \cite{tian2019distributed} whose vertices
correspond to robots, and edges represent the presence of at least one inter-robot loop closure between the two corresponding robots (Fig.~\ref{fig:robust_init}).
Note that the spanning tree induces a unique path between any two robots.
Without loss of generality, we select an arbitrary robot $\alpha$ and use its reference frame $A$ as the global frame.
For each remaining robot $\beta$, we need to obtain its relative transformation to the global frame $\widehat{\MX}^A_B \in \SEthree$. 
This is done by traversing the unique path in the robot-level spanning tree from $\alpha$ to $\beta$, and composing all estimated \emph{pairwise} transformations computed using \eqref{eq:pose_avg} along the way.
In practice, this procedure can be performed in a fully distributed fashion, 
by incrementally growing the robot-level spanning tree from $\alpha$ using local communication.
Finally, each robot $\beta$ uses its corresponding $\widehat{\MX}^A_B$ to express its initial trajectory in the global frame.
Note that our distributed PGO approach does not require the robots to share these initial trajectory estimates, but only requires them to be expressed in a shared global frame at each robot.

\subsection{Robust Distributed Pose-Graph Optimization}
\label{sec:dpgo}

Following the initialization stage, robots perform robust distributed PGO to obtain optimal trajectory estimates while simultaneously rejecting outlier loop closures.
Let $\MX_{\alpha_i} = (\MR_{\alpha_i}, \vt_{\alpha_i}) \in \SEthree$ denote the $i$th pose of robot $\alpha$ in the global frame.
We aim to optimize all pose variables using all 
odometric measurements and putative loop closures.
\begin{equation}
\begin{aligned}
\min_{
	\substack{
	\MX_{\alpha_i} \in \SEthree, \\
	\forall \alpha \in \robot, \; \forall i}} \quad 
& 
\underbrace{
\sum_{\alpha \in \robot} \sum_{i=1}^{n_\alpha-1} 
r_{\alpha_i}(\MX_{\alpha_i},\MX_{\alpha_{i+1}})^2}_{\text{odometry}} 
+ \\
&
\underbrace{
\sum_{(\alpha_i,\beta_j) \in L} 
\rho \big (r^{\alpha_i}_{\beta_i}(\MX_{\alpha_i},\MX_{\beta_{j}}) \big),}_{\text{loop closures}} 
\end{aligned}
\label{eq:robust_pgo}
\end{equation}
where $\robot = \{\alpha, \beta, \hdots\}$ denotes the set of robots,
$n_\alpha$ is the total number of poses of robot $\alpha$, and the set of loop closures $L$ includes both intra-robot and inter-robot loop closures.
Each residual error in \eqref{eq:robust_pgo} corresponds to a single relative pose measurement in the global pose graph, where the residual error is measured using the chordal distance. For example, the residual corresponding to a loop closure is given by~\cite{Rosen18ijrr-sesync}, 
\begin{equation}
\begin{aligned} 
	r^{\alpha_i}_{\beta_i}(\MX_{\alpha_i},\MX_{\beta_{j}}) \triangleq
	\bigg (
	& w_R \norm{\MR_{\beta_j} - \MR_{\alpha_i} \widetilde{\MR}^{\alpha_i}_{\beta_j}}^2_F
	+ \\
	& w_t \norm{\vt_{\beta_j} - \vt_{\alpha_i} - \MR_{\alpha_i} \widetilde{\vt}^{\alpha_i}_{\beta_j}}^2_2
	\bigg)^{1/2},
\end{aligned}
\end{equation} 
where $\widetilde{\MX}^{\alpha_i}_{\beta_j} = (\widetilde{\MR}^{\alpha_i}_{\beta_j}, \widetilde{\vt}^{\alpha_i}_{\beta_j}) \in \SEthree$ is the observed noisy transformation, and $w_R, w_t >0$ specify measurement precisions.
We employ the standard quadratic cost for odometric measurements as they are outlier-free.
For loop closures, we choose $\rho$ to be the TLS function as in Section~\ref{sec:init}.

To solve \eqref{eq:robust_pgo}, we develop a \emph{fully distributed} variant of GNC which uses the Riemannian block-coordinate descent (\RBCD) solver \cite{tian2019distributed} as the workhorse during iterative optimization.
Recall from Section~\ref{sec:gnc} that GNC alternates between 
variable (\ie trajectory) updates and weight updates.
In the following, we discuss how each of these two operations are performed in the distributed setup.

\myParagraph{Variable update}
In this case, the variable update step becomes an instance of standard (weighted) PGO,
\begin{equation}
	\begin{aligned}
		\min_{
			\substack{
			\MX_{\alpha_i} \in \SEthree, \\
			\forall \alpha \in \robot, \; \forall i}
		} \quad 
		& 
		\sum_{\alpha \in \robot} \sum_{i=1}^{n_\alpha-1} 
		r_{\alpha_i}(\MX_{\alpha_i},\MX_{\alpha_{i+1}})^2
		+ \\
		&
		\sum_{(\alpha_i,\beta_j) \in L} 
		w^{\alpha_i}_{\beta_j} \cdot r^{\alpha_i}_{\beta_i}(\MX_{\alpha_i},\MX_{\beta_{j}})^2.
	\end{aligned}
	\label{eq:pgo_var_update}
\end{equation}
Compared to \eqref{eq:robust_pgo}, terms including the robust cost function $\rho$ (corresponding to the loop closures) are replaced by weighted squared residuals; see also~\eqref{eq:var_update}.
We apply the \RBCD solver \cite{tian2019distributed} for distributed optimization
of \eqref{eq:pgo_var_update} (Algorithm~\ref{alg:dgnc}, line~\ref{alg:dgnc:var_update}).
In short, RBCD operates on the rank-restricted relaxation~\cite{Rosen18ijrr-sesync} of \eqref{eq:pgo_var_update} and subsequently projects the solution to the Special Euclidean group.
In our implementation, we set the default rank relaxation to 5.
\RBCD is a fully decentralized algorithm, in which each robot $\alpha \in \robot$ is responsible for estimating its own trajectory $\MX_{\alpha} \triangleq \{\MX_{\alpha_i},  i = 1, \hdots, n_\alpha\}$.
During execution, robots alternate to update their trajectories by relying on partial information exchange with their teammates.
Specifically, at each iteration in which robot $\alpha$ updates its trajectory, it needs to communicate once with its neighboring robots (\ie robots that share inter-robot loop closures with robot $\alpha$), where the communication can be either direct or relayed by other robots.
Furthermore, robot $\alpha$ only needs to receive neighboring robots' ``public poses''
(\ie poses that share inter-robot loop closures with robot $\alpha$).
This property allows \RBCD to preserve privacy and saves communication effort over the remaining poses.
The main advantages of RBCD over the previous DGS method \cite{Choudhary17ijrr-distributedPGO3D} lies in the fact that it has provable convergence guarantees.
Moreover, RBCD can be used as an anytime algorithm, since each iteration is guaranteed to improve over the previous iterates by reducing the PGO cost function, while DGS requires completing rotation estimation before initiating pose estimation.
We refer interested readers to \cite{tian2019distributed} for complete details about \RBCD.

In the original (centralized) GNC algorithm, each variable update step is solved to full convergence using a global solver or local search technique.
In the distributed setup, however, solving each instance of \eqref{eq:pgo_var_update} to full convergence 
can be slow, due to the first-order nature of typical distributed optimization methods (including both \RBCD and DGS).
To develop a more practical and efficient approach, we relax the convergence requirements and allow \emph{approximate} solutions during variable updates.
Specifically, we only apply \RBCD for a fixed number of iterations to refine the trajectory estimates based on the current weights.
In our implementation, we set the number of iterations to 15 by default.
The resulting trajectories are then used to warm start the next variable update step.
As measurement weights converge, our approach also allows the trajectory estimates to converge to relatively high precision.

\myParagraph{Weight update}
In the original GNC paper \citep{Yang20ral-GNC}, it has been shown that the weight update for each residual function using TLS only depends on the current residual error $\widehat{r}_i$, control parameter $\mu$, and the threshold $\bar{c}$ of the TLS cost,
\begin{equation}
	w_i \leftarrow \begin{cases}
	0, 
	& \text{ if } \widehat{r}^2_i \in \big[\frac{\mu+1}{\mu}\bar{c}^2,  +\infty \big], \\
	\frac{\bar{c}}{\widehat{r}_i} \sqrt{\mu(\mu+1)} - \mu, 
	& \text{ if } \widehat{r}^2_i \in \big [ \frac{\mu}{\mu + 1}\bar{c}^2, \frac{\mu+1}{\mu}\bar{c}^2 \big], \\
	1, 
	& \text{ if } \widehat{r}^2_i \in \big [0, \frac{\mu}{\mu+1} \bar{c}^2].
	\end{cases}
	\label{eq:pgo_weight_update}
\end{equation}
See \cite[Proposition~4]{Yang20ral-GNC} for more details.
The weight update step is particularly suitable for distributed computation, 
as \eqref{eq:pgo_weight_update} suggests that this operation can be performed 
\emph{independently and in parallel} for each residual function (\ie loop closure).
We leverage this insight to implement a fully distributed weight update scheme (Algorithm~\ref{alg:dgnc}, line~\ref{alg:dgnc:weight_update}).
Specifically, each robot first updates weights associated with its internal loop closures in parallel.
Then, for each inter-robot loop closure, one of the two involved robots computes the updated weight, and subsequently transmits the new weight to the other robot.
After the weight update stage, each robot also updates its local copy of the control parameter $\mu$, so that the sequence of surrogate cost functions gradually converges to the original TLS function (Algorithm~\ref{alg:dgnc}, line~\ref{alg:dgnc:mu_update}).


\section{Local Mesh Optimization}
\label{sec:pgmo}

This section describes how to perform local correction of the 3D mesh in response to a loop closure.
\kimeraSemantics builds the 3D mesh from the \kimeraVIO (odometric) estimate.
However, since distributed PGO described in previous section improves the accuracy of the trajectory estimate by enforcing loop closures, it is desirable 
to correct the mesh according to the optimized trajectory estimate (\ie each time distributed PGO is executed). 
Here we propose an approach for mesh optimization based on deformation graphs \cite{Summer07siggraph-embeddedDeformation}. 
\emph{Deformation graphs} are a model from computer graphics 
that deforms a given mesh in order to anchor points in this mesh to user-defined locations while ensuring that the 
mesh remains locally rigid; deformation graphs are typically used for 3D animations, where one wants to animate a 3D object while 
ensuring it moves smoothly and without artifacts~\cite{Summer07siggraph-embeddedDeformation}.

\myParagraph{Creating the deformation graph}
In our approach, we create a unified deformation graph including a simplified mesh and a pose graph of trajectory keyframes.
The process is illustrated in Fig.~\ref{fig:deformation-graph}.
The intuition is that the "anchor points" in~\cite{Summer07siggraph-embeddedDeformation} will be the keyframes in our trajectory.
More specifically, while \kimeraSemantics builds a local 3D mesh for each robot $\alpha$ using pose estimates from \kimeraVIO, 
we keep track of the subset of 3D mesh vertices seen in each  keyframe from \kimeraVIO. 
To build the deformation graph, 
we first subsample the mesh from \kimeraSemantics to obtain a simplified mesh. 
We simplify the mesh with an online vertex clustering method by storing the vertices of the mesh in an octree data structure;
as the mesh grows, the vertices in the same voxel of the octree are merged and degenerate faces and edges are removed. 
The voxel size is tuned according to the environment or the dataset. 
Then, the vertices of this simplified mesh and the corresponding keyframe poses 
are added as \emph{vertices} in the deformation graph; we are going to refer to the corresponding vertices in the 
deformation graph as \emph{mesh vertices} and \emph{keyframe vertices}.
Moreover, we add two types of \emph{edges} to the deformation graph:
\emph{mesh edges} (corresponding to pairs of mesh vertices sharing a face in the simplified mesh),  
and \emph{keyframe edges} (connecting a keyframe with the set of mesh vertices it observes).

For each mesh vertex $k$ in the deformation graph, we assign a transformation
$\MM_k = (\MR^M_k,\vt^M_k)$, where $\MR^M_k  \in \SOthree$  and  $\vt^M_k \in \Real{3}$; 
$\MM_k$ defines a local coordinate frame, where
$\MR_k$ is initialized to the identity and $\vt_k$ is initialized to the position $\vg_k$ of the mesh vertex 
from \kimeraSemantics (\ie without accounting for loop closures). 
We also assign a pose $\MX_i = (\MR^x_k,\vt^x_k)$ to each keyframe vertex $i$.
The pose is initialized to the pose estimates from \kimeraVIO.

\begin{figure}
\centering
\subfloat[Undeformed mesh.]{\includegraphics[width=0.5\columnwidth]{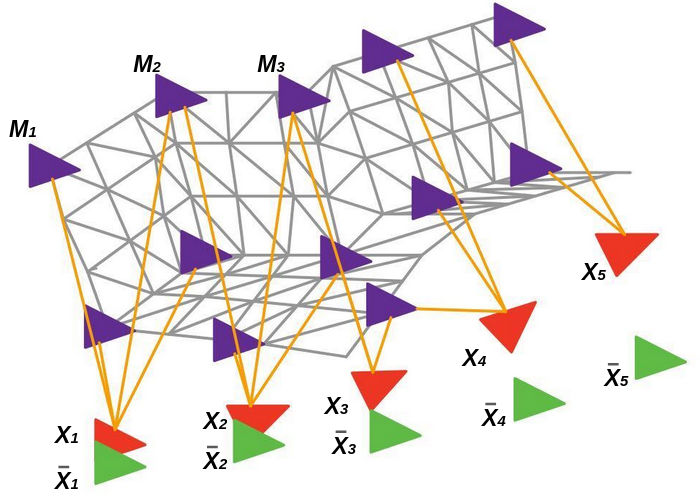}}
\subfloat[Deformed mesh]{\includegraphics[width=0.5\columnwidth]{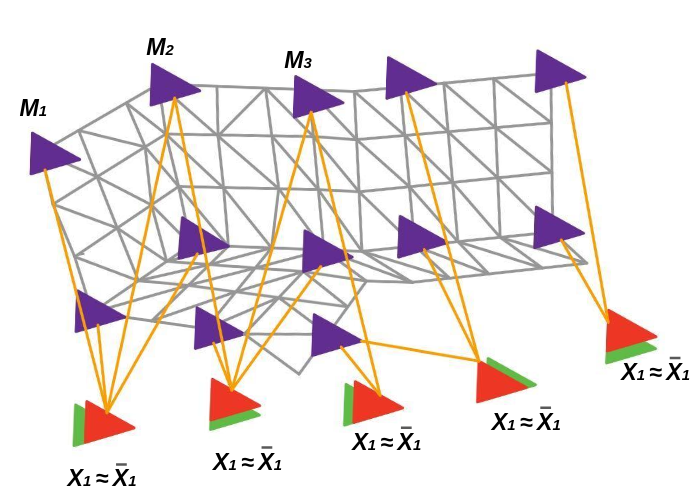}}
\caption{LMO deformation graph including mesh vertices (violet) and keyframe vertices (red). Edges connect two mesh vertices that 
are adjacent in the mesh (gray links), as well as mesh vertices with the keyframe vertices they are observed in (orange links).
The green poses denote optimized poses from distributed PGO.}
\label{fig:deformation-graph}
\vspace{-6mm}
\end{figure}

\myParagraph{Optimizing the deformation graph}
The goal is to correct the 
mesh on each robot in response to changes in the keyframe poses (due to PGO).
Towards this goal, we need to adjust the poses (and the mesh vertex positions) to ``anchor'' 
the keyframe poses to the latest estimates from distributed PGO 
as shown in Fig.~\ref{fig:deformation-graph}.
Denoting the optimized poses from distributed PGO as $\bar{\MX}_i$, 
and calling $n$ the number of keyframes in the trajectory and $m$ the total number of mesh vertices in the deformation graph.
Following \citep{Summer07siggraph-embeddedDeformation},
we compute updated poses $\MX_i$, $\MM_k$ of the vertices in the deformation graph 
by solving the following \emph{local} optimization problem at each robot,

\vspace{-3mm}
\begin{align}
\label{eq:lmo}
    \argmin_{\substack{\MX_1,\ldots,\MX_n \in \SEthree \\ 
    \MM_1, \ldots,\MM_m \in \SEthree}} &
    \sum_{i=0}^n|| \MX_i \boxminus \bar{\MX}_i ||^2_{\Sigma_x} + \notag\\
  &\sum_{k=0}^m\sum_{l \in \mathcal{N}^M(k)}||\MR^M_k(\vg_l - \vg_k) + \vt^M_k - \vt^M_l||^2_{\Sigma} + \notag\\ 
  &\sum_{i=0}^n\sum_{l \in \mathcal{N}^M(i)}||\MR^x_i\widetilde{\vg}_{il} + \vt^x_i - \vt^M_l||^2_{\Sigma}
  \vspace{-3mm}
\end{align}
where $\vg_k$ denotes the non-deformed position of vertex $k$ in the deformation graph, 
$\widetilde{\vg}_{il}$ denotes the non-deformed position of vertex $l$ in the coordinate frame of keyframe $i$, 
 $\mathcal{N}^M(k)$ 
 denotes all the mesh vertices in the deformation graph connected to vertex $k$, 
 and $\boxminus$ denotes a tangent-space representation of the relative pose between  $\MX_i$ and $ \bar{\MX}_i$~\cite[7.1]{Barfoot17book}. 
Intuitively, the first term in the minimization~\eqref{eq:lmo} enforces (``anchors'') the poses of each keyframe $\MX_i$ 
to match the optimized poses $\bar{\MX}_i$ from distributed PGO.  
The second term enforces local rigidity of the mesh by minimizing the mismatch with respect to the non-deformed configuration
$\vg_k$.
The third term enforces local rigidity of  the relative positions between keyframes and mesh vertices by
minimizing the mismatch with respect to the non-deformed configuration
in the local frame of pose $\MX_i$.
We optimize~\eqref{eq:lmo} using a Levenberg-Marquardt method in GTSAM~\cite{gtsam}.

Since the deformation graph contains a subsampled version of the original mesh,
after the optimization, we retrieve the location of the remaining vertices as in~\cite{Summer07siggraph-embeddedDeformation}.
In particular, the positions of the vertices of the complete mesh are obtained 
as affine transformations of nodes in the deformation graph: 
\vspace{-1mm}
\begin{equation}
  \widetilde{\vv}_i = \sum_{j=1}^m s_j(\vv_i)[\MR^M_j(\vv_i - \vg_j) + \vt^M_j]
  \vspace{-1mm}
\end{equation}
where $\vv_i$ indicates the original vertex positions and $\widetilde{\vv}_i$ are the new deformed positions. 
The weights $s_j$ are defined as
\vspace{-1mm}
\begin{equation}
  s_j(\vv_i) = \left(1 - ||\vv_i - \vg_j||/{d_{\text{max}}}\right)^2
  \vspace{-1mm}
\end{equation}
and then normalized to sum to one. Here $d_{\text{max}}$ is the distance to the $k + 1$ nearest node 
as described in \cite{Summer07siggraph-embeddedDeformation} (we set $k=4$).

Note that the \kimeraSemantics mesh also includes semantic labels as an attribute for each node in the mesh, which remain untouched in the mesh deformation.

\section{Experiments}
\label{sec:experiments}

In this section, we perform extensive evaluations of \kimeraMulti.
Our results show that \kimeraMulti provides robust and accurate estimation of trajectories and metric-semantic meshes, is efficient in terms of communication usage, and is flexible thanks to its modularity.
The rest of this section is organized as follows.
In Section~\ref{sec:robustness_experiment}, 
we analyze the robustness of \kimeraMulti in numerical experiments.
In Section~\ref{sec:dataset_evaluation}, we evaluate the quality of trajectory estimates and metric-semantic reconstruction in photo-realistic simulations and benchmarking datasets.
Lastly, in Section~\ref{sec:outdoor_datasets}, we demonstrate \kimeraMulti on two challenging real-world datasets collected by ground robots.

\subsection{PGO Robustness Analysis}
\label{sec:robustness_experiment}
In this section, we evaluate different robust trajectory estimation techniques on synthetic datasets with varying ratios of outlier loop closures.
Our results demonstrate the importance of robust initialization for multi-robot PGO.
Furthermore, we show that alternative technique based on PCM~\cite{Mangelson18icra} has low recall (\ie missing correct loop closures).
Overall, we show that the proposed \DGNC method achieves the best performance, and is not sensitive to parameter tuning.

\myParagraph{Single-robot experiments}
To offer additional insights and contrast with the multi-robot analysis later, we first perform ablation studies on single-robot synthetic datasets.
We simulate 2D PGO problems contaminated by outliers using the \scenario{INTEL} dataset~\citep{Carlone14tro-SO2}.
To generate outlier loop closures, we randomly select pairs of non-adjacent poses in the original pose graph, and add relative measurements with uniformly random rotations and translations. For translations, we sample each coordinate uniformly at random within the domain $[-10,10]$~m.

The following trajectory estimation techniques are compared:
(1) {L2}: standard least squares optimization using Levenberg-Marquardt (LM),
(2) {PCM}: outlier rejection with pairwise consistency maximization \cite{Mangelson18icra} using the approximate maximum clique solver \citep{Rossi15parallel} followed by LM, 
(3) {GNC}: graduated non-convexity~\cite{Yang20ral-GNC},
(4) {PCM + GNC}: PCM outlier rejection followed by GNC. 
Both LM and GNC are implemented in GTSAM~\cite{gtsam}.
All methods start from the odometry initial guess.
Note that both PCM and GNC require the user to specify a confidence level (in the form of a probability threshold) that determines the maximum residual of inliers. 
We vary this probability threshold and compare different techniques across the entire spectrum.

\begin{figure}[t]
	\centering
	\subfloat[Outlier ratio: 10\%]{%
		\includegraphics[trim=5 0 30 30, clip, width=0.24\textwidth]
		{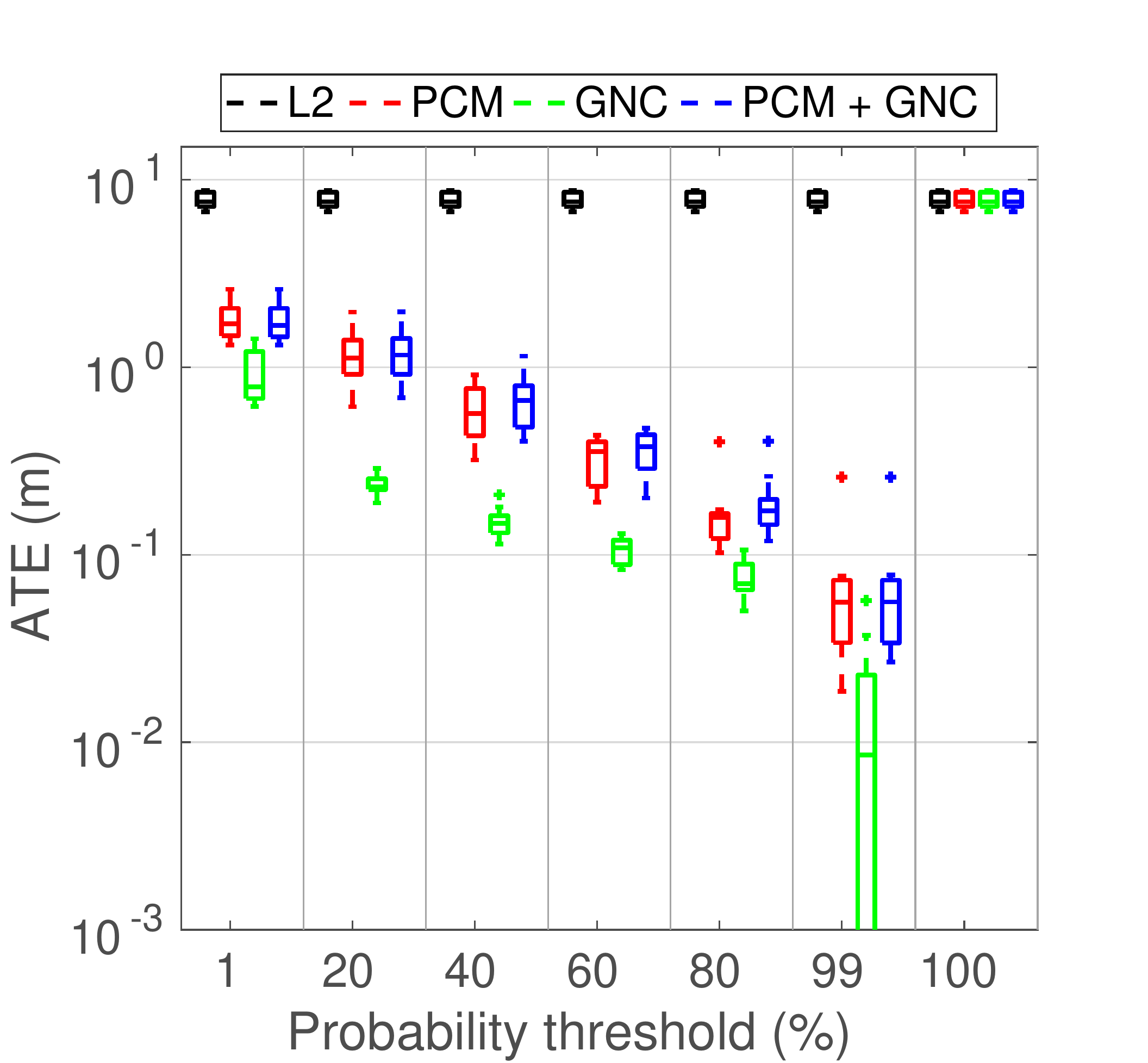}
		\label{fig:robustness_single_robot_p1}
	} ~
	\subfloat[Outlier ratio: 70\%]{%
		\includegraphics[trim=5 0 30 30, clip, width=0.24\textwidth]
		{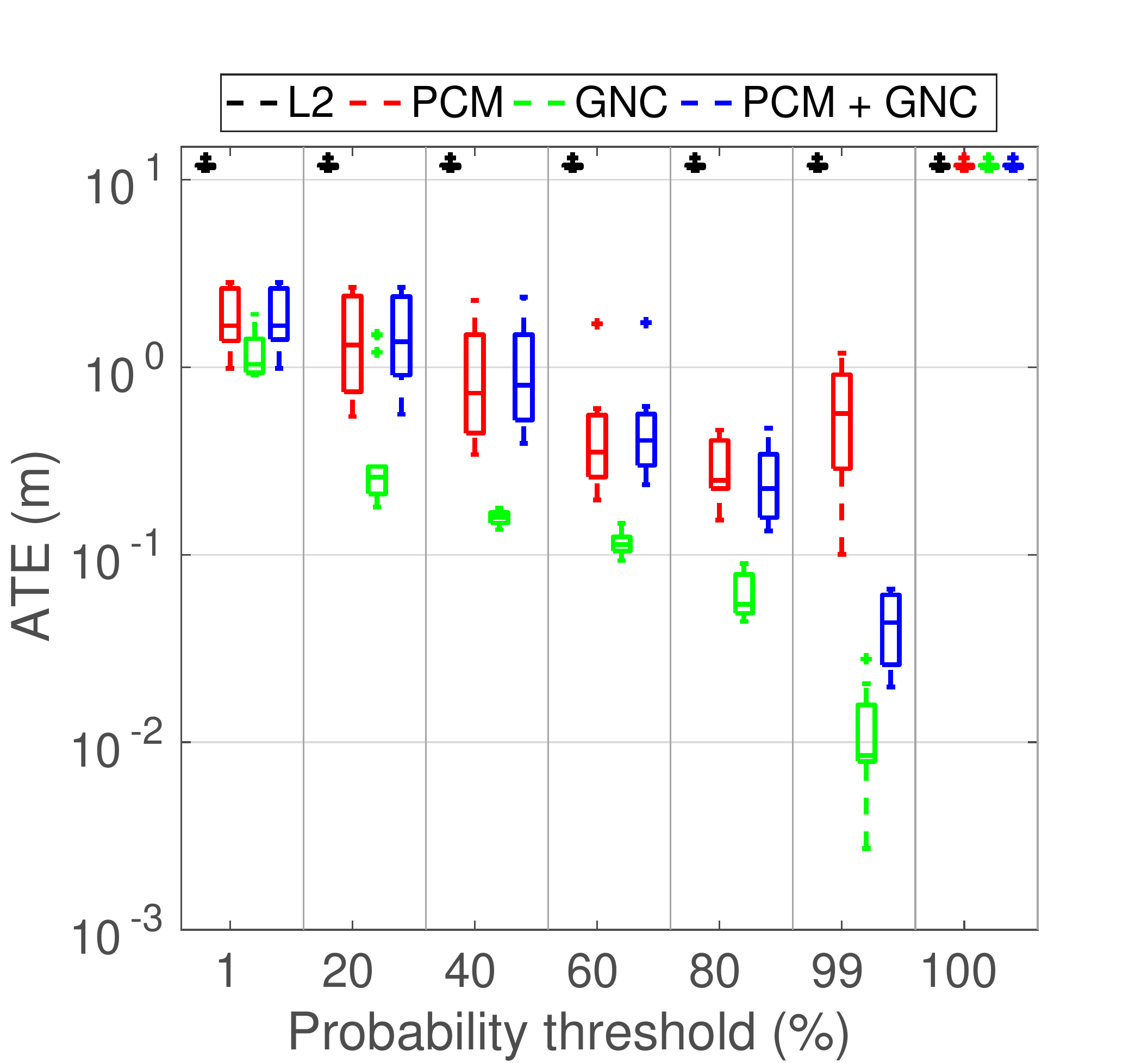}
		\label{fig:robustness_single_robot_p7}
	}
	\caption{\textbf{Single-robot tests.} Comparisons between solvers on single-robot synthetic PGO problems across 10 Monte Carlo runs.
	} \label{fig:robustness_single_robot}
\end{figure}

Fig.~\ref{fig:robustness_single_robot} shows the absolute trajectory errors (ATE) with respect to the maximum likelihood estimate, computed using the outlier-free pose graph. Results are collected over 10 Monte Carlo runs. 
Standard LM optimization is not robust even under 10\% outlier loop closures
(Fig.~\ref{fig:robustness_single_robot_p1}).
In many cases, PCM tends to be overly conservative and reject inliers (due to approximate maximum clique search), which leads to an increase in the trajectory error.
The same issue also negatively impacts the performance of PCM + GNC (blue), since rejected inliers cannot be recovered. 
On the other hand, GNC (green) achieves smaller error across the entire spectrum.
Under 70\% outliers (Fig.~\ref{fig:robustness_single_robot_p7}), 
PCM has larger errors especially at higher probability thresholds (\eg 99\%), 
indicating that the method is unable to reject all outliers. 
In this case, applying subsequent GNC helps to improve the performance of PCM.
However, also in this case,
applying GNC alone consistently achieves the best performance over the entire range of probability thresholds.
This result suggests that GNC should be the method of choice in single-robot PGO independent from the parameter tuning.

\myParagraph{Multi-robot experiments}
In multi-robot PGO, there is no longer an outlier-free initial guess (\ie odometry), which is crucial for the strong performance of GNC observed in the single-robot case.
We investigate this issue in the next experiment, and demonstrate the robust initialization scheme proposed in Section~\ref{sec:init} as an effective solution.
Similar to the previous experiment, we use the \scenario{INTEL} dataset with the same outlier model described previously.
The pose graph is divided into three segments with approximately equal lengths to simulate a three-robot collaborative SLAM scenario.

We compare two variants of GNC using different initial guesses.
The first variant uses the proposed robust initialization scheme,
and is labeled as ``GNC'' in Fig.~\ref{fig:robustness_multi_robot} (green). 
The second variant uses a na\"ive initialization formed using the local odometry of each robot and randomly sampled inter-robot loop closures between pairs of robots; see \eqref{eq:naive_inter_frame}.
This variant is labeled as ``GNC (na\"ive init)'' in Fig.~\ref{fig:robustness_multi_robot} (magenta).
When PCM is used, we sample inter-robot loop closures from the inlier set returned by PCM.
All problems are solved using a centralized implementation based on GTSAM \citep{gtsam}.
Distributed experiments will be presented in the next section. 

\begin{figure}[t]
	\centering
	\subfloat[Outlier ratio: 10\%]{%
		\includegraphics[trim=5 0 30 30, clip, width=0.24\textwidth]
		{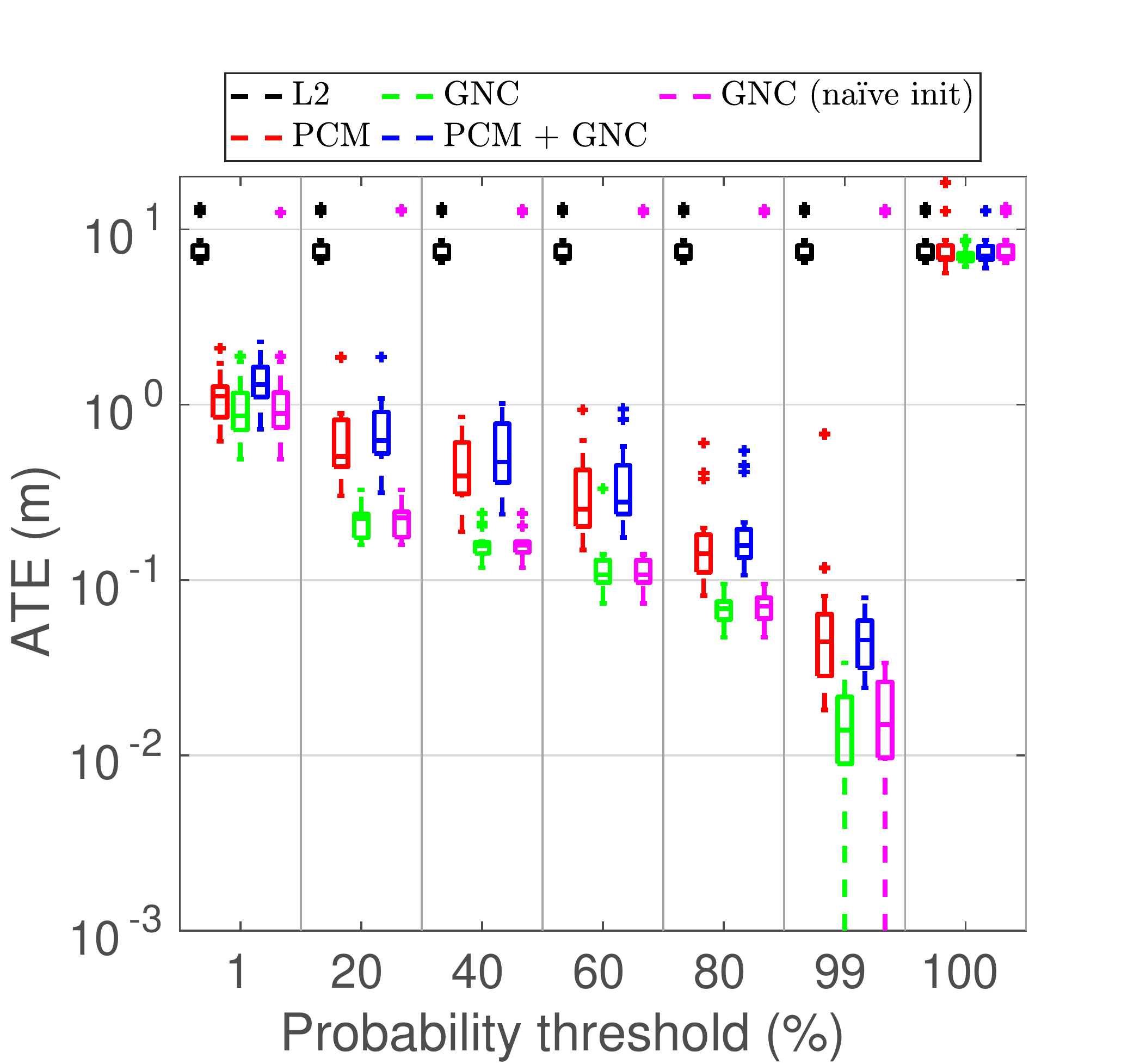}
		\label{fig:robustness_multi_robot_p1}
	} ~
	\subfloat[Outlier ratio: 70\%]{%
		\includegraphics[trim=5 0 30 30, clip, width=0.24\textwidth]
		{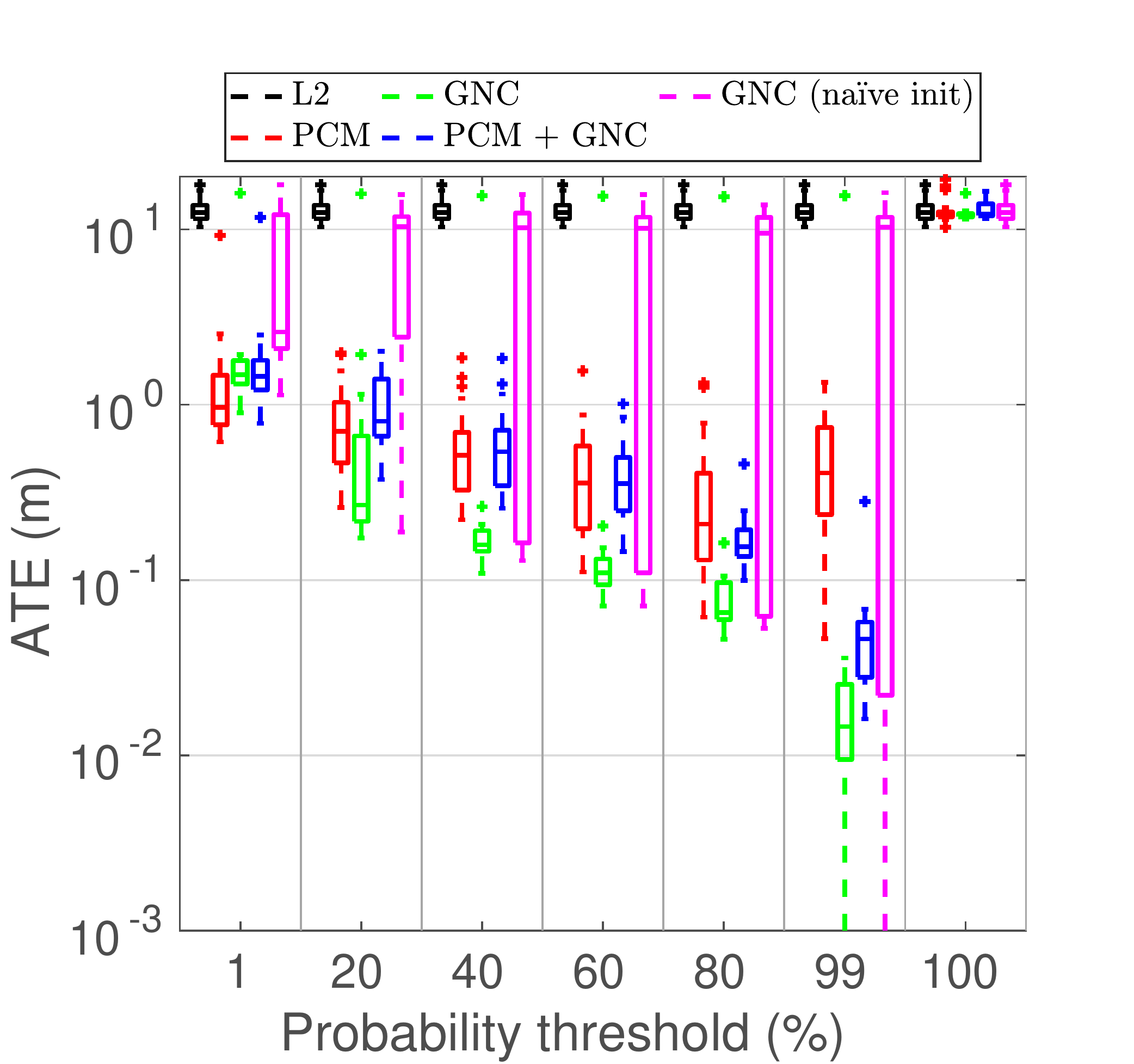}
		\label{fig:robustness_multi_robot_p7}
	}
	\caption{
		\textbf{Multi-robot tests.}
		Comparisons between solvers on three-robot synthetic PGO problems across 10 Monte Carlo runs.
		}
	\label{fig:robustness_multi_robot}
\end{figure}

\begin{figure*}[t]
	\centering
	\subfloat[{PCM} (ATE = 2.24m)]{%
		\includegraphics[trim=30 0 30 35, clip, width=0.24\textwidth]
		{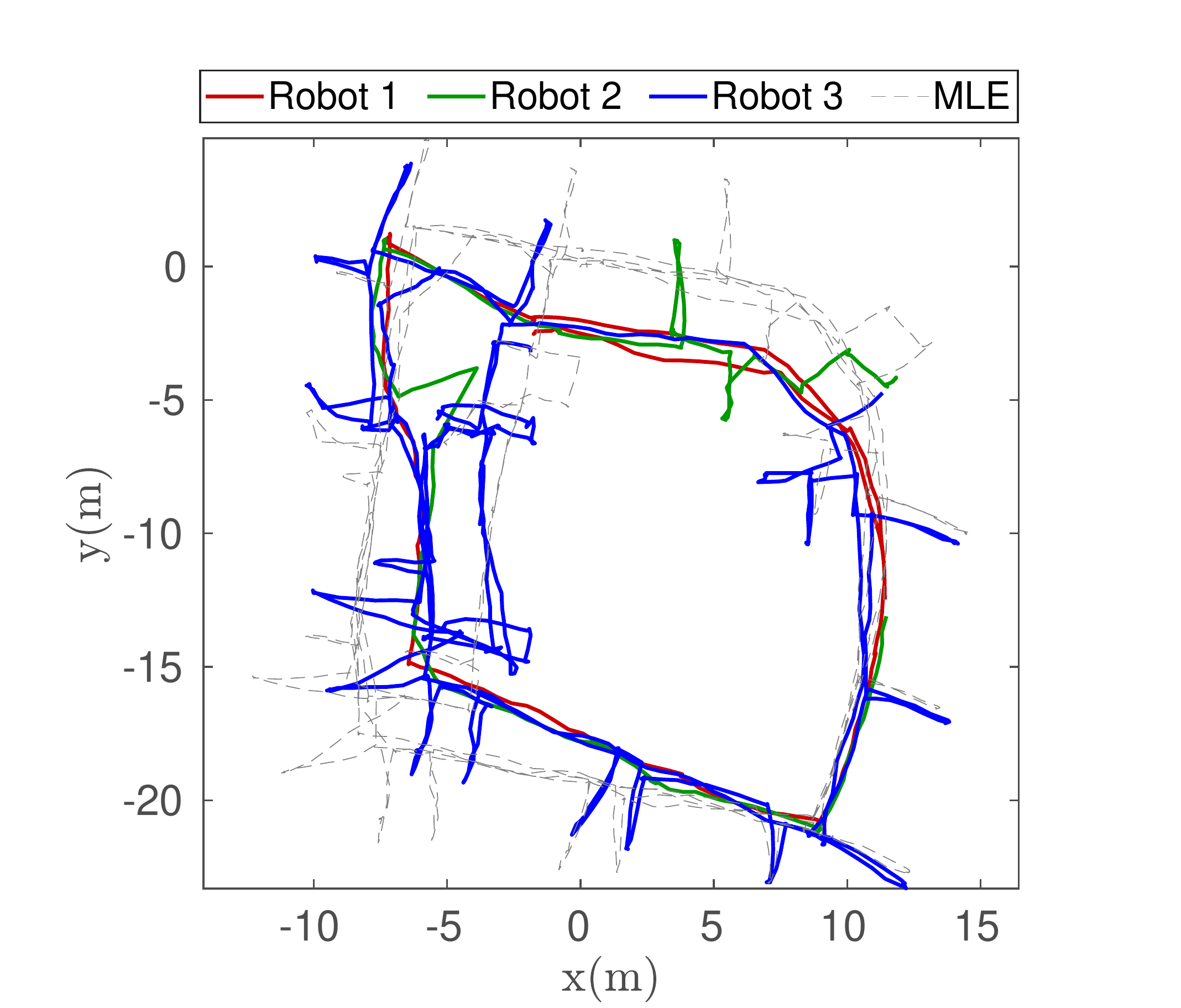}
		\label{fig:robustness_multi_robot_p7_pcm}
	} ~
	\subfloat[{GNC (na\"ive init)} (ATE = 11.59m)]{%
		\includegraphics[trim=30 0 30 35, clip, width=0.24\textwidth]
		{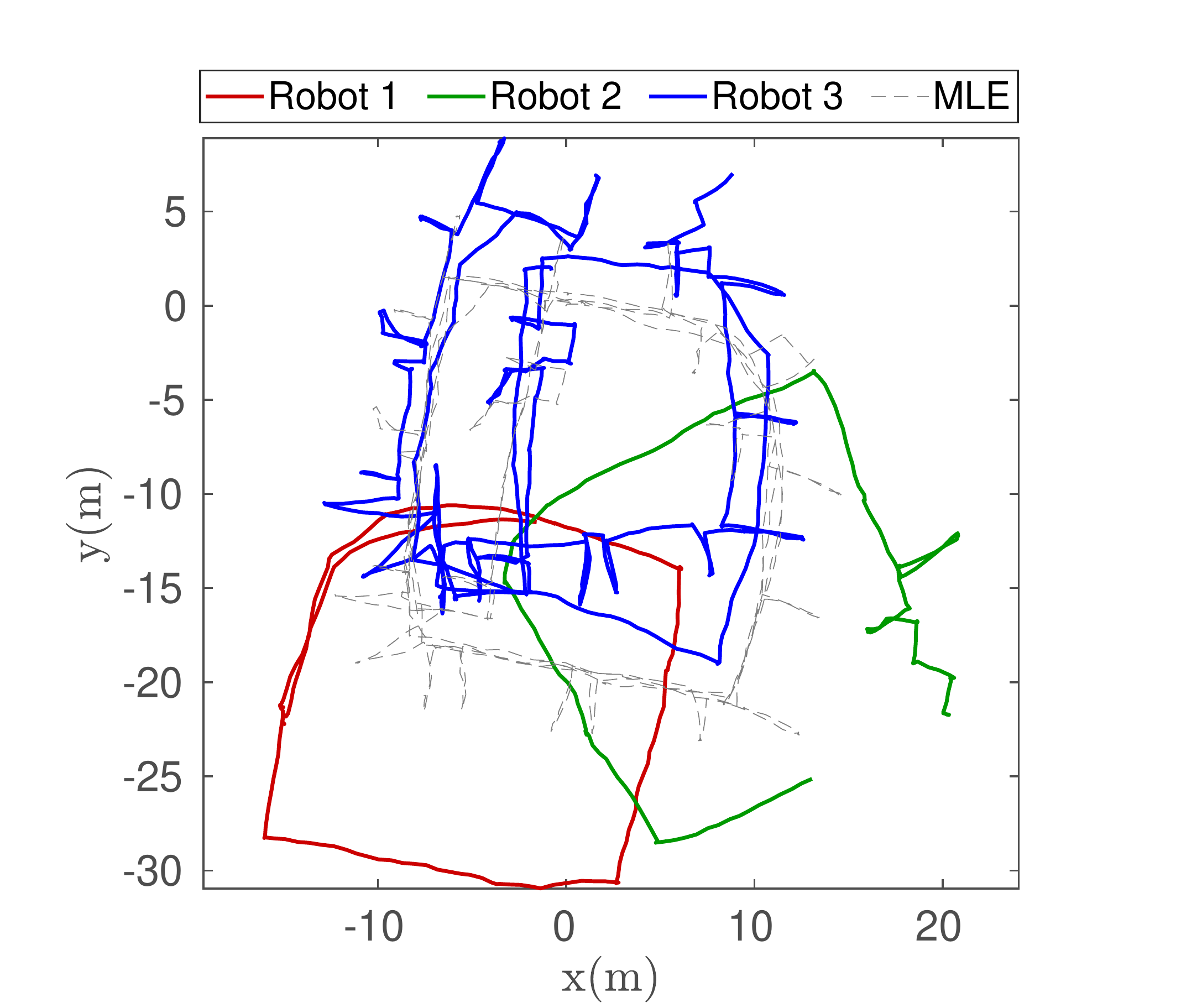}
		\label{fig:robustness_multi_robot_p7_gnc}
	}~
	\subfloat[{PCM + GNC} (ATE = 0.09m)]{%
		\includegraphics[trim=30 0 30 35, clip, width=0.24\textwidth]
		{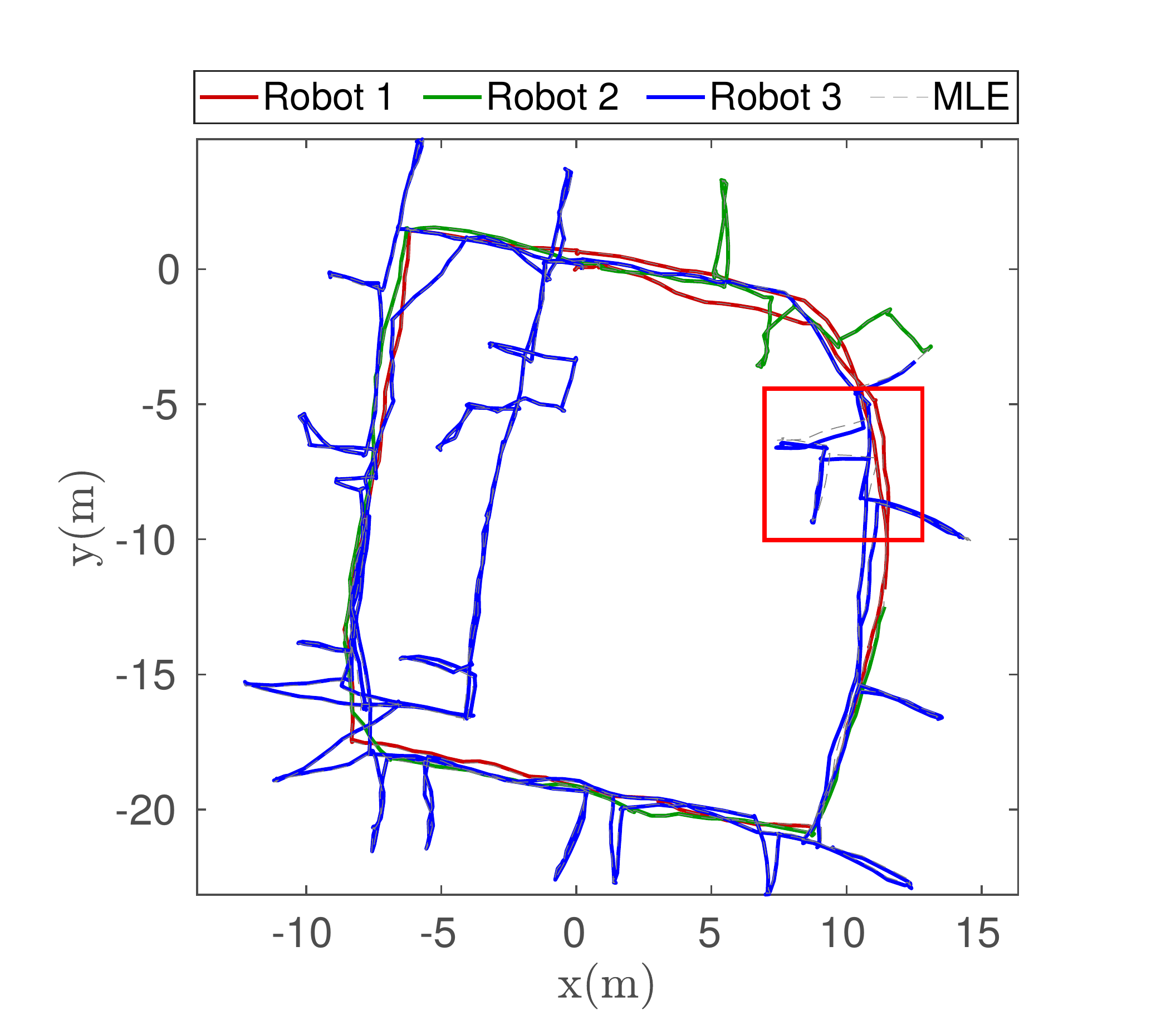}
		\label{fig:robustness_multi_robot_p7_both}
	}
	~
	\subfloat[{GNC}  (ATE = 0.003m)]{%
		\includegraphics[trim=30 0 30 35, clip, width=0.24\textwidth]
		{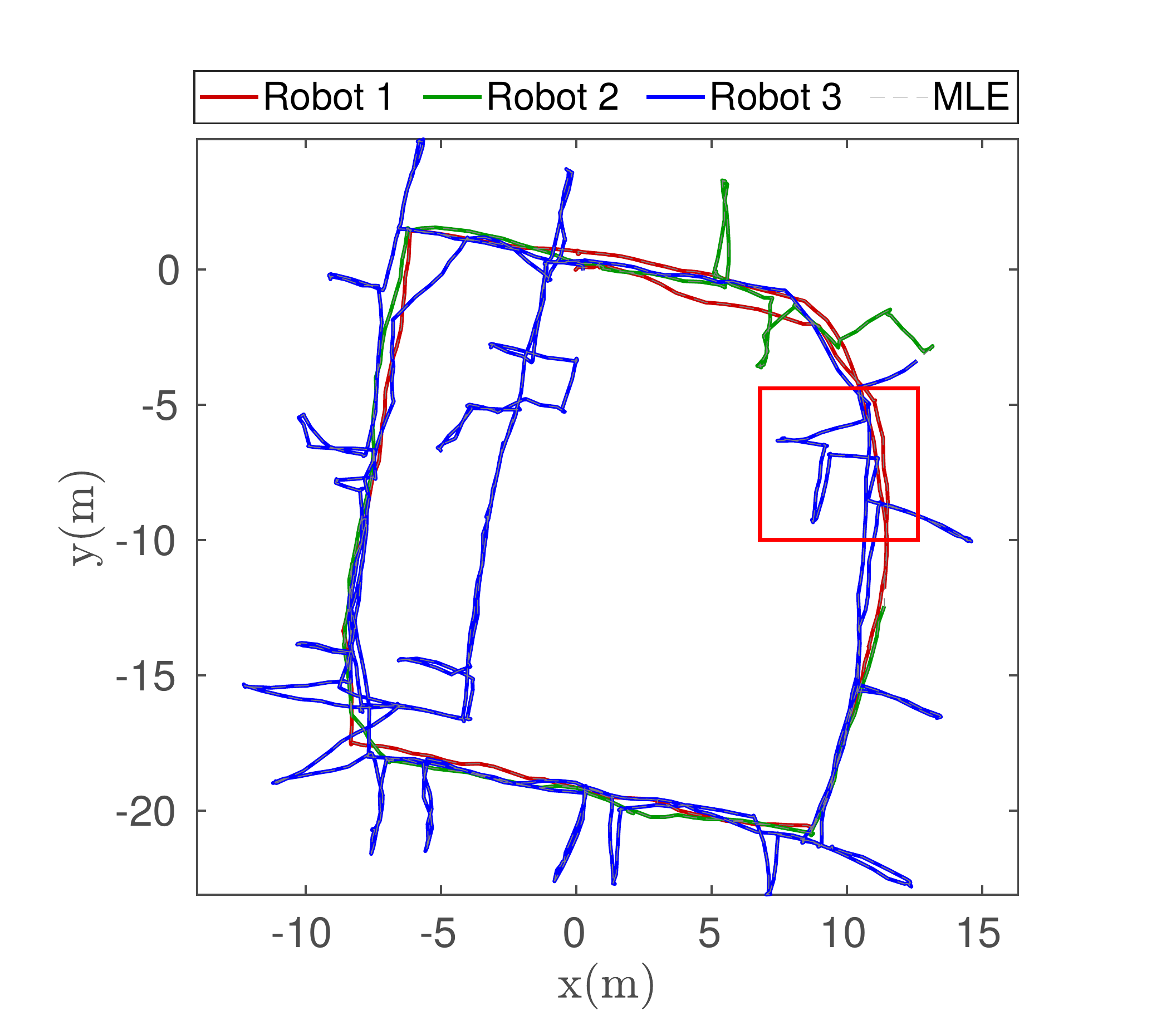}
		\label{fig:robustness_multi_robot_p7_gncalign}
	}
	\caption{Comparing final trajectory estimates of different techniques under 70\% outlier loop closures. All methods use the same probability threshold of 99\%.}
	\label{fig:robustness_multi_robot_traj}
\end{figure*}

Fig.~\ref{fig:robustness_multi_robot} reports ATE results across 10 Monte Carlo runs.
With 10\% outlier loop closures (Fig.~\ref{fig:robustness_multi_robot_p1}), it is less likely that the na\"ive initialization is affected by outliers.
Consequently, the two variants of GNC have similar performance in most cases, but na\"ive initialization still causes occasional failures (magenta outliers).
The failure cases correspond to instances when the initial guess was accidentally built using an outlier loop closure.
The problem caused by incorrect initialization becomes more evident under 70\% outlier loop closures (Fig.~\ref{fig:robustness_multi_robot_p7}), where na\"ive initialization fails in the majority of instances.
This is because under 70\% outliers, the na\"ive initial guess is almost always contaminated by wrong loop closures, which severely affects the performance of GNC.
In comparison, 
using PCM helps to avoid catastrophic failures, but PCM still exhibits low recall as in the single-robot case.
Finally, the proposed robust initialization effectively corrects the wrong initial guess, and applying GNC from the robust initialization (green) consistently outperforms other techniques.

To provide additional insights over the performance of different techniques,
Fig.~\ref{fig:robustness_multi_robot_traj} shows qualitative comparisons of final trajectory estimates on a random problem instance with 70\% outliers. All techniques use the same probability threshold of 99\%.
Under this setting, PCM (Fig.~\ref{fig:robustness_multi_robot_p7_pcm}) fails to reject all outlier loop closures.
As a result, its solution is distorted when compared to the maximum likelihood estimate.
When applying GNC from na\"ive initialization (Fig.~\ref{fig:robustness_multi_robot_p7_gnc}), 
the method fails to recover any inlier loop closures
due to incorrect initialization that causes the variable update to converge to wrong estimates.
\mbox{Fig.~\ref{fig:robustness_multi_robot_p7_both}-\ref{fig:robustness_multi_robot_p7_gncalign}} show that
applying either PCM or robust initialization
to correct the initial guess before applying GNC can effectively resolve the problem.
Between this two approaches, however, our proposed robust initialization
produces lower trajectory error, which can also be seen by comparing the trajectory estimates within the red box.
This is because PCM incorrectly removes inlier loop closures during outlier rejection, which causes a loss of accuracy that cannot be recovered by GNC.

\subsection{Evaluation in Simulation and Benchmarking Datasets}
\label{sec:dataset_evaluation}

\begin{table*}[t]
	\centering
	\caption{Absolute trajectory errors (ATE) in meters with respect to ground truth trajectories.
	For each dataset, we also report the total trajectory length (including all robots).
	L2: standard least squares optimization using LM;
	PCM: pairwise consistency maximization~\citep{Mangelson18icra};
	\DGNC: proposed distributed trajectory estimation method (using robust initialization);
	NI: na\"ive initialization;
	ES: early stopping.
	For reference, we also report the ATE of centralized GNC 
	(colored in {\color[HTML]{656565} \bfseries gray}).
	}
	\label{tab:dataset_ate}
	\setlength{\tabcolsep}{8.0pt}
	\renewcommand{\arraystretch}{1.5}
	\begin{tabular}{|c|c|c|c|c|c|c|c|c|}
		\hline
                       & Length [m] & L2 & PCM & \DGNC (NI) & PCM + \DGNC & \DGNC & \DGNC(ES) & Centralized GNC \\ \hline
        Medfield   & 2396 & 64.2 & 12.5 & 57.4 & 4.64 & \textbf{3.92} & 4.32 &
        {\color[HTML]{656565} \bfseries 3.88} \\ \hline               
		City       & 1213 & 3.58 & 1.57 & 0.91 & 1.08 & 0.85 & \textbf{0.76} & 
		{\color[HTML]{656565} \bfseries 1.00} \\ \hline
		Camp       & 1037 & 11.9 & 1.37 & 0.97 & 1.09 & 0.96 & \textbf{0.75} & 
		{\color[HTML]{656565} \bfseries 1.33} \\ \hline
		Vicon Room 1   & 211 & 1.17 & 1.00 & 0.34 & 0.45 & 0.35 & \textbf{0.21} & 
		{\color[HTML]{656565} \bfseries 0.36} \\ \hline
		Vicon Room 2   & 206 & 1.87 & 1.56 & 0.46 & 0.62 & \textbf{0.47} & 0.48 & 
		{\color[HTML]{656565} \bfseries 0.43} \\ \hline
		Machine Hall   & 466 & 1.92 & 1.76 & 0.48 & 0.70 & \textbf{0.41} & 0.49 & 
		{\color[HTML]{656565} \bfseries 0.52} \\ \hline
	\end{tabular}%
\end{table*}

We evaluate \kimeraMulti in three photo-realistic simulation environments 
(\medfield, \city, \camp), developed by the Army Research Laboratory Distributed and Collaborative Intelligent Systems and Technology (DCIST) Collaborative Research Alliance~\citep{dcist}.
In addition, we also evaluate on three real-world environments
(\scenario{Vicon Room 1, Vicon Room 2, Machine Hall}) from the \euroc dataset~\citep{Burri16ijrr-eurocDataset}.
Among all datasets, \scenario{Machine Hall} contains five sequences which are used to simulate collaborative SLAM with five robots.
The simulation and \scenario{Vicon Room} datasets contain three sequences that are used to simulate a three-robot scenario. 
In our experiments in this section and Section~\ref{sec:outdoor_datasets},
we run \kimeraMulti in a setting where robots are constantly in communication range,
which means that inter-robot loop closures are established at the earliest possible time.
In future work, we plan to further improve our implementation
and test our system in scenarios where communication is intermittent.

\myParagraph{Trajectory estimation results}
We first evaluate the accuracy of different distributed trajectory estimation techniques.
In this experiment, we use \kimeraVIO to process raw sensor data, and it is thus hard to obtain accurate covariance information for all measurements.
In our implementation, we use a fixed isotropic covariance for each residual in PGO, with a standard deviation of 0.01~rad for rotation and 
0.1~m for translation.
Moreover, we use a relatively conservative probability threshold of 50\% for all robust estimation techniques.
We compare the following distributed solvers:
(1) L2: standard PGO (least squares optimization) using \RBCD \citep{tian2019distributed},
(2) PCM: outlier rejection with PCM~\cite{Mangelson18icra}, followed by \RBCD,
(3) \DGNC (NI): proposed \DGNC method starting from a na\"ive initial guess that combines local odometry of each robot with randomly sampled inter-robot loop closures between pairs of robots,
(4) PCM + \DGNC: outlier rejection with PCM, followed by \DGNC,
(5) \DGNC: the proposed \DGNC method with robust initialization, 
(6) \DGNC (ES): an ``early stopped'' version of \DGNC that terminates after 50 total \RBCD updates,
(7) centralized GNC from GTSAM~\cite{gtsam}.

Table~\ref{tab:dataset_ate} reports the final ATE of each method when evaluated against the ground truth. 
Note that the total trajectory length varies significantly across datasets, 
which also causes ATE to vary.
Due to the existence of outlier loop closures, 
standard least squares optimization (L2) gives large errors.
PCM improves over the L2 results, but still yields large errors on a subset of datasets.
The proposed \DGNC method achieves significantly lower trajectory errors on all datasets.
Similar to the synthetic experiments (Section~\ref{sec:robustness_experiment}), we observe that applying GNC after 
PCM (``PCM + \DGNC'' in the table) always leads to suboptimal performance compared to the proposed approach, due to the low recall of PCM.
On the \medfield simulation, applying \DGNC from na\"ive initialization fails.
In this case, the na\"ive initialization is wrong due to the selection of an outlier loop closure.
This creates an error in the initial alignment of robots' reference frames which \DGNC is unable to correct.
Finally, we observe that on three of the datasets, applying early stopping (ES) leads to lower error compared to full optimization (distributed or centralized).
In this experiment, estimation errors are computed with respect to the ground truth trajectories, which are in general different from the true (unknown) maximum likelihood estimate.
In summary, the proposed \DGNC method achieves the best performance, and applying early stopping (ES) does not significantly affect the accuracy of trajectory estimation, which remains comparable to the centralized GNC.

\begin{table*}[t]
	\centering
	\caption{ 
		Communication usage and solution runtime.
		The data payloads induced by \kimeraMulti are further divided into 
		three modules: 
		place recognition (PR) that exchanges bag-of-word vectors,
		geometric verification (GV) that transmits keypoints and feature descriptors,
		and distributed pose graph optimization (DPGO).
		Centralized communication and runtime are
		colored in {\color[HTML]{656565} \bfseries gray}.
	}
	\label{tab:dataset_comm}
	\setlength{\tabcolsep}{4pt}
	\renewcommand{\arraystretch}{1.5}
	\begin{tabular}{|c|c|c|c|c|c|c|c|c|c|c|c|}
		\hline
		\multirow{2}{*}{Dataset} & \multirow{2}{*}{\# Poses} & \multirow{2}{*}{\# Edges} & \multicolumn{6}{c|}{Communication {[}MB{]}}    & \multicolumn{3}{c|}{Runtime [sec]} \\ \cline{4-12} 
		&
		&
		&
		PR &
		GV &
		DPGO &
		Total &
		\begin{tabular}[c]{@{}c@{}}Centralized\\ (Images)\end{tabular} &
		\begin{tabular}[c]{@{}c@{}}Centralized\\ (Keypoints)\end{tabular} &
		\begin{tabular}[c]{@{}c@{}}Distributed  \end{tabular} &
		\begin{tabular}[c]{@{}c@{}}Distributed (ES) \end{tabular} &
		\begin{tabular}[c]{@{}c@{}}Centralized \end{tabular} \\ \hline
		Medfield             & 2918   & 3104   & 22.6 & 41.5 & 1.8  & \textbf{65.9} & {\color[HTML]{656565} \bfseries 2113} & {\color[HTML]{656565} \bfseries 141}  
		& 29.2  & \textbf{5.9}  & {\color[HTML]{656565} \bfseries 4.4} \\ \hline
		City                 & 3212   & 4173   & 16.2 & 44.5 & 8.8  & \textbf{69.5} & {\color[HTML]{656565} \bfseries 2326} & {\color[HTML]{656565} \bfseries 155}   
		& 22.1  & \textbf{4.5}  & {\color[HTML]{656565} \bfseries 3.2} \\ \hline
		Camp                 & 5088   & 5200   & 39.2 & 19.7 & 0.5  & \textbf{59.4} & {\color[HTML]{656565} \bfseries 3685} & {\color[HTML]{656565} \bfseries 246}   
		& 43.2  & \textbf{9.1}  & {\color[HTML]{656565} \bfseries 4.4} \\ \hline
		Vicon Room 1             & 1693   & 2788   & 9.5  & 14.7 & 3.6  & \textbf{27.8} & {\color[HTML]{656565} \bfseries 1226} & {\color[HTML]{656565} \bfseries 81.7}  
		& 8.9   & \textbf{2.2}  & {\color[HTML]{656565} \bfseries 3.1} \\ \hline
		Vicon Room 2             & 1738   & 2335   & 11.5 & 10.2 & 2.7  & \textbf{24.4} & {\color[HTML]{656565} \bfseries 1259} & {\color[HTML]{656565} \bfseries 83.9}  
		& 11.7  & \textbf{3.2}  & {\color[HTML]{656565} \bfseries 1.7} \\ \hline
		Machine Hall             & 3261   & 5196   & 46.0 & 76.8 & 22.9 & \textbf{145.7}& {\color[HTML]{656565} \bfseries 2362} & {\color[HTML]{656565} \bfseries 157}   
		& 20.5  & \textbf{2.5}  & {\color[HTML]{656565} \bfseries 6.3} \\ \hline
	\end{tabular}
\end{table*}

\myParagraph{Communication usage and solution runtime}
In Table~\ref{tab:dataset_comm}, we compare the communication usage of \kimeraMulti with two baseline centralized architectures that either transmit all images or keypoints.
Data payloads used by \kimeraMulti are divided into three parts: place recognition (exchanging bag-of-word vectors), geometric verification (exchanging keypoints and descriptors), and distributed PGO.
The front-end (first two modules) consumes more communication than the back-end (distributed PGO).
Overall, our results demonstrate that \kimeraMulti is communication-efficient.
For instance, on the \scenario{Vicon Room 2} dataset, our system achieves a communication reduction of 70\% compared to the baseline centralized system that transmits all keypoints and descriptors.
On the other hand, the system does not achieve equally significant communication reduction on the \scenario{Machine Hall} dataset. 
Compared to other datasets, the increased number of robots in \scenario{Machine Hall} results in more data transmission.
In particular, the loose thresholds for loop closure detection lead to increased data transmission during the geometric verification (GV) stage.
Further communication reduction may be achieved by employing recent 
communication-efficient methods for distributed place recognition~\cite{Cieslewski17Netvlad, Cieslewski18icra}
and geometric verification~\cite{tian2018near,tian2019resource}.

In addition, Table~\ref{tab:dataset_comm} reports the runtime of \DGNC
and also compares with the centralized solver (implemented in GTSAM~\citep{gtsam}).
Our method has reasonable runtime (approximately 10 seconds) for the smaller \scenario{Vicon Room} datasets.
For the larger datasets, \DGNC requires more time for full convergence. 
Nevertheless, applying early stopping (ES) effectively keeps the runtime close to its centralized counterpart, without heavily compromising estimation accuracy.

\myParagraph{Metric-Semantic Mesh Quality}
We use the ground-truth point clouds available in the \euroc Vicon Room 1 and 2 datasets, and the ground-truth mesh (and its semantic labels) available 
in the DCIST simulator to evaluate the accuracy of the 3D metric-semantic mesh built by \kimeraSemantics and the impact of the local mesh optimization (LMO). 
For evaluation, the estimated and ground-truth meshes are sampled with a uniform density of $10^3~\text{points}/\text{m}^2$ as in~\cite{Rosinol20icra-Kimera}.
The resulting semantically-labeled point clouds are then registered using the ICP~\cite{Besl92pami} implementation
in \emph{Open3D}~\cite{Zhou18arxiv-open3D}. 
Then, we calculate the mean distance between each point in the ground-truth point cloud to its nearest neighbor in the estimated point cloud to obtain the metric accuracy of the 3D mesh. 
In addition, we evaluate the semantic reconstruction accuracy 
by calculating the percentage of correctly labeled points~\cite{Rosinol20icra-Kimera} relative to the ground truth 
using the correspondences given by ICP.
Fig.~\ref{fig:euroc_metric_accuracy} and Fig.~\ref{fig:sim_metric_accuracy} report the metric accuracy  
of the individual meshes constructed by each robot as well as the merged 
global mesh, 
and Table~\ref{tab:semantic_eval_sim} shows the semantic reconstruction accuracy 
in the simulator (\euroc does not provide ground-truth semantics). In general, the metric-semantic mesh accuracy improves 
after LMO for both individual and merged 3D meshes, 
demonstrating the effectiveness of LMO in conjunction with our distributed trajectory optimization.
The dense metric-semantic meshes are shown in Fig.~\ref{fig:metric_semantic_reconstruction_camp} and Fig.~\ref{fig:metric_semantic_reconstruction_city}. In the case when semantic labels are unavailable, we are still able to 
generate the mesh, colored by the RGB image colors, as shown in Fig.~\ref{fig:metric_semantic_reconstruction_medfield} 
for the experiment in the simulator portraying the \scenario{Medfield} scene. 

\begin{figure}[t]
	\vspace{-2mm}
	\centering
	\includegraphics[
		trim=0 0 0 0, clip,
		width=0.49\textwidth]
		{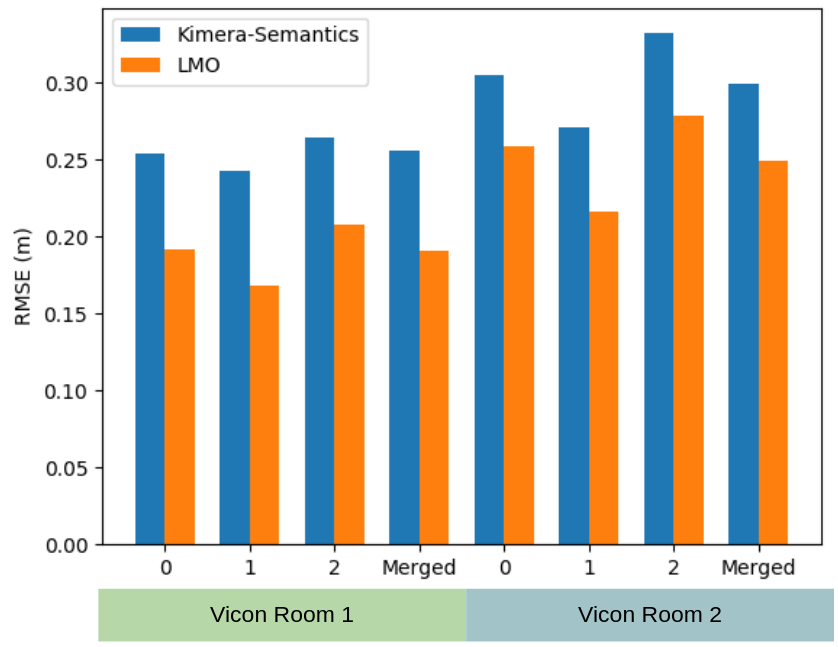}
	\vspace{-1mm}
	\caption{Metric reconstruction evaluation on the Euroc squences. Mesh error (in meters) for the 3D meshes by \kimeraSemantics and \kimeraMulti's LMO.
	\label{fig:euroc_metric_accuracy} \vspace{-1mm}}
\end{figure}

\begin{figure}[t]
	\vspace{-2mm}
	\centering
	\includegraphics[
		trim=0 0 0 0, clip,
		width=0.49\textwidth]
		{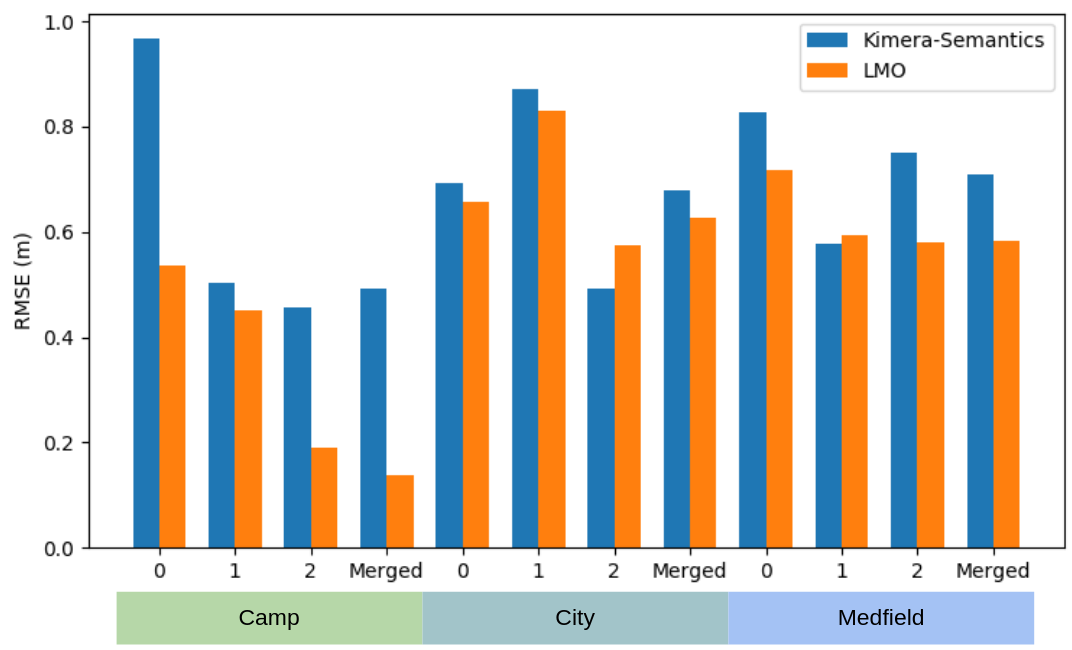}
	\vspace{-1mm}
	\caption{Metric reconstruction evaluation on the Camp, City, and Medfield simulator datasets. Mesh error (in meters) for the 3D meshes by \kimeraSemantics and \kimeraMulti's LMO.
	\label{fig:sim_metric_accuracy} \vspace{-1mm}}
\end{figure}

\begin{table}[t]
	\caption{
		\footnotesize
		Semantic reconstruction evaluation. Semantic labels accuracy before and after correction by LMO in the DCIST simulator. \vspace{-2mm} }
	\label{tab:semantic_eval_sim}
	\centering
	\setlength{\tabcolsep}{1.6pt}
	\renewcommand{\arraystretch}{1.2}
	\begin{tabular}{|c|c|c|c|}
		\hline
		Dataset       & Robot ID & \kimeraSemantics (\%) &     LMO (\%)     \\ \hline
		& 0       &    81.6             &    \bf{96.2}    \\
		Camp          & 1       &    92.8        &    \bf{98.1}         \\ 
		& 2       &    82.8             &    \bf{96.1}    \\ 
		& Merged   &    79.4             &    \bf{95.2}    \\
		\hline
		& 0       &    77.1        &    \bf{77.7}         \\
		City          & 1       &    80.7             &    \bf{83.1}    \\ 
		& 2       &    \bf{71.4}             &    70.6    \\
		& Merged   &    76.1             &    \bf{78.8}    \\ 
		\hline
		
	\end{tabular}
	\vspace{-3mm}
\end{table}

\begin{figure}[t]
	\centering
	\includegraphics[width=0.99\columnwidth,
	trim=0mm 0mm 0mm 0mm,clip]{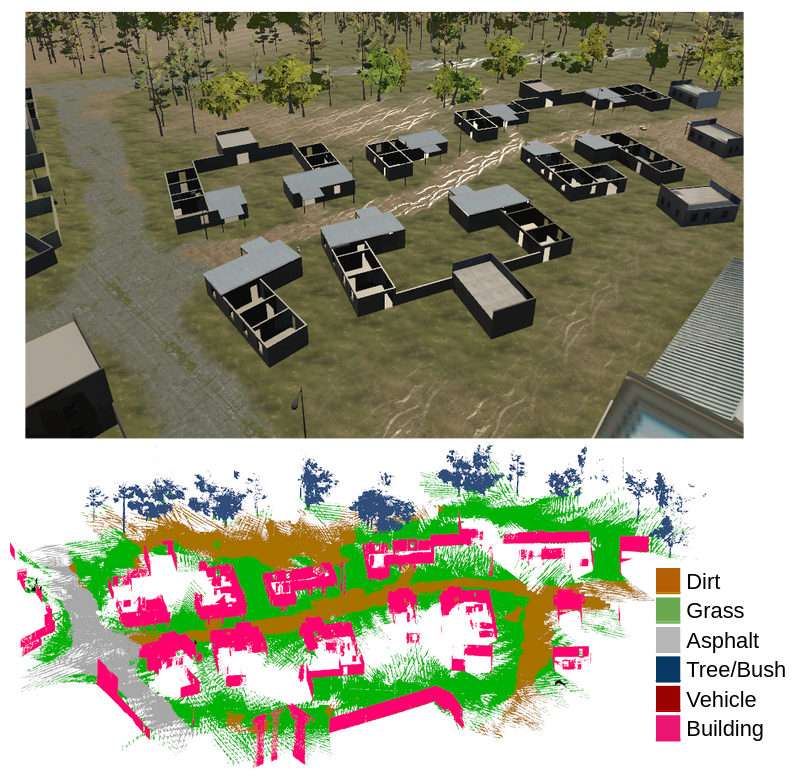}
	\vspace{-6mm}
	\caption{
		Dense metric-semantic 3D mesh model generated by Kimera-Multi with three robots in the simulated \scenario{Camp} scene. \label{fig:metric_semantic_reconstruction_camp} }
	\vspace{-1mm}
\end{figure}

\begin{figure}[t]
	\centering
	\includegraphics[width=0.99\columnwidth,
	trim=0mm 0mm 0mm 0mm,clip]{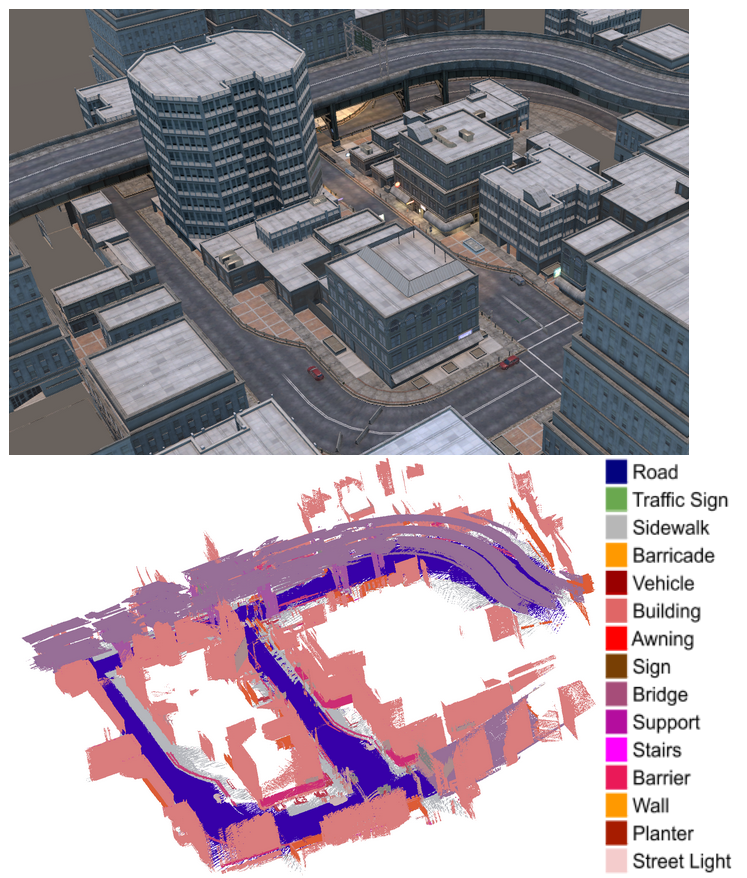}
	\vspace{-6mm}
	\caption{
		Dense metric-semantic 3D mesh model generated by Kimera-Multi with three robots in the simulated \scenario{City} scene. \label{fig:metric_semantic_reconstruction_city} }
	\vspace{-1mm}
\end{figure}

\begin{figure}[t]
	\centering
	\includegraphics[width=0.99\columnwidth,
	trim=0mm 0mm 0mm 0mm,clip]{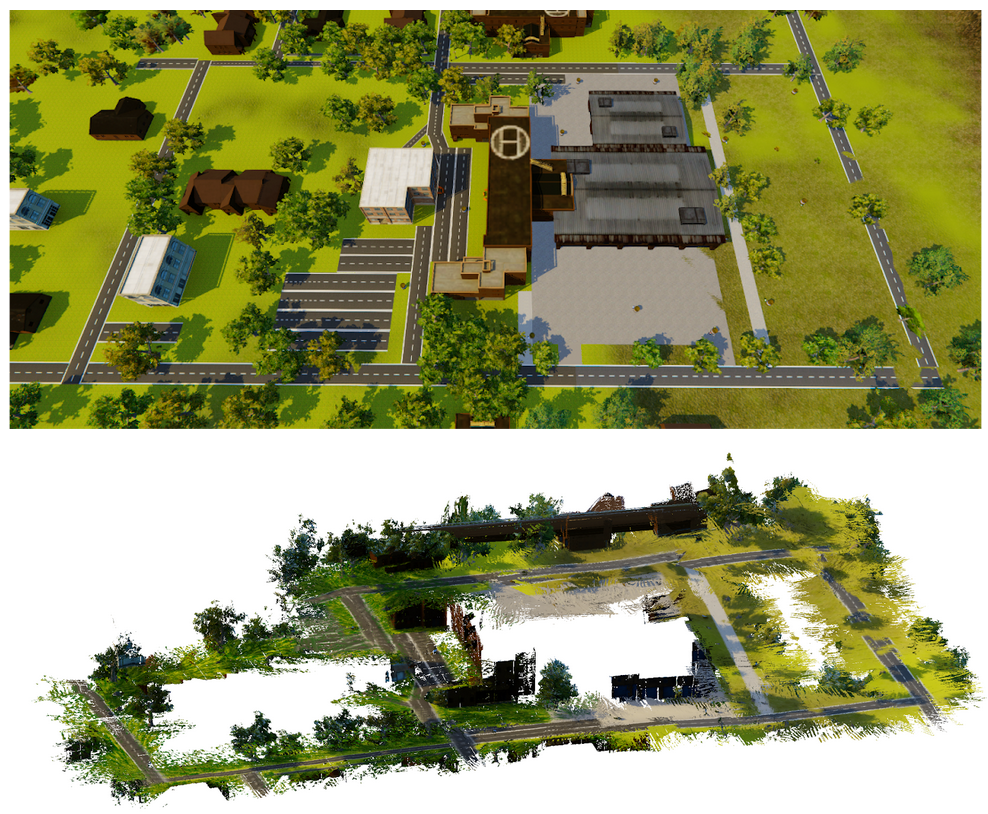}
	\vspace{-6mm}
	\caption{
		Dense metric 3D mesh model generated by Kimera-Multi with three robots in the simulated Medfield scene. \label{fig:metric_semantic_reconstruction_medfield} }
	\vspace{-1mm}
\end{figure}

\subsection{Evaluation in Large-Scale Outdoor Datasets}
\label{sec:outdoor_datasets}

\myParagraph{Experimental Setup}
We demonstrate \kimeraMulti on two challenging outdoor datasets,
collected using a Clearpath Jackal UGV equipped with a forward-facing RealSense D435i RGBD Camera and IMU. 
The first dataset was collected at the Medfield State Hospital, Massachusetts, USA (Fig.~\ref{fig:real-world-experiment:medfield}).
Three sets of trajectories were recorded, with the longest trajectory being 860 meters in length. 
The second dataset was collected around the Ray and Maria Stata Center at MIT (Fig.~\ref{fig:real-world-experiment:stata}), and also includes three different trajectories with each trajectory being over 500 meters in length. 
In both Fig.~\ref{fig:real-world-experiment:medfield} and Fig.~\ref{fig:real-world-experiment:stata},
the red, orange, and blue trajectories correspond to robots with ID 0, 1, and 2, respectively.
Both sets of experiments are challenging and include many similar-looking scenes that induce spurious loop closures.

\myParagraph{Results and Discussions}
Table~\ref{tab:real-world-loop-closures} reports statistics about loop closures on the outdoor datasets.
Specifically, for each pair of robots, we report the number of loop closures accepted by \DGNC over the total number of detected loop closures (including outliers).
Diagonal entries in the table correspond to intra-robot loop closures.
Both datasets contain many outlier loop closures, which are successfully rejected by \DGNC.
Compared to \medfield, the \Stata dataset contains significantly less inter-robot loop closures, which makes distributed PGO particularly challenging.

In order to evaluate estimation accuracy in the absence of ground truth trajectories, we measure end-to-end errors as in \citep{Forster17tro}.
In particular, we design each individual robot trajectory to start and finish at the same place, and then compute the final end-to-end position errors.
The end-to-end error is not equivalent to the ATE, but still provides useful information about the final estimation drift on each trajectory.
Table~\ref{tab:real-world-end-to-end} compares the end-to-end errors of \kimeraVIO, 
\kimeraMulti (using \DGNC to estimate trajectories), and centralized result (solved using GNC in GTSAM~\citep{gtsam}).
To complement the quantitative result, we also provide qualitative visualizations of the optimized trajectories and meshes in Fig.~\ref{fig:real-world-experiment:medfield} and Fig.~\ref{fig:real-world-experiment:stata}. 

\begin{figure*}[th]
	\centering
	\subfloat[\kimeraVIO]{%
		\includegraphics[width=0.45\textwidth]{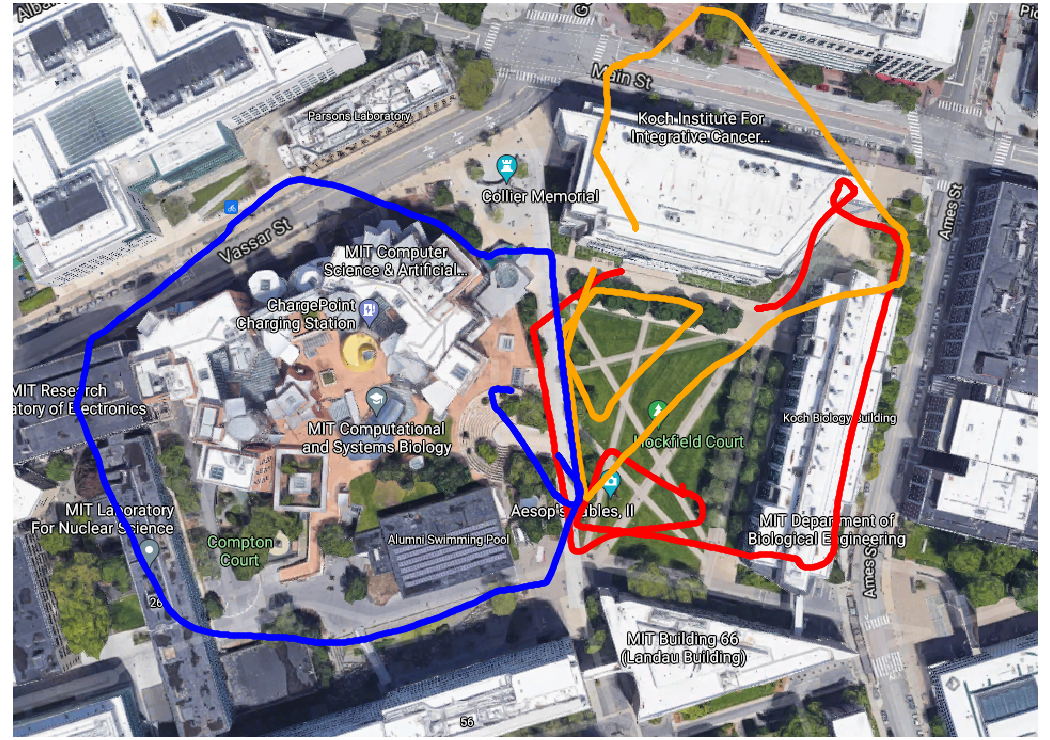}
		\label{fig:stata_vio_traj}
	}
	\hspace{23pt}
	\subfloat[\kimeraMulti (\DGNC with approximate variable updates)]{%
		\includegraphics[trim=345 120 345 100, clip, width=0.41\textwidth]{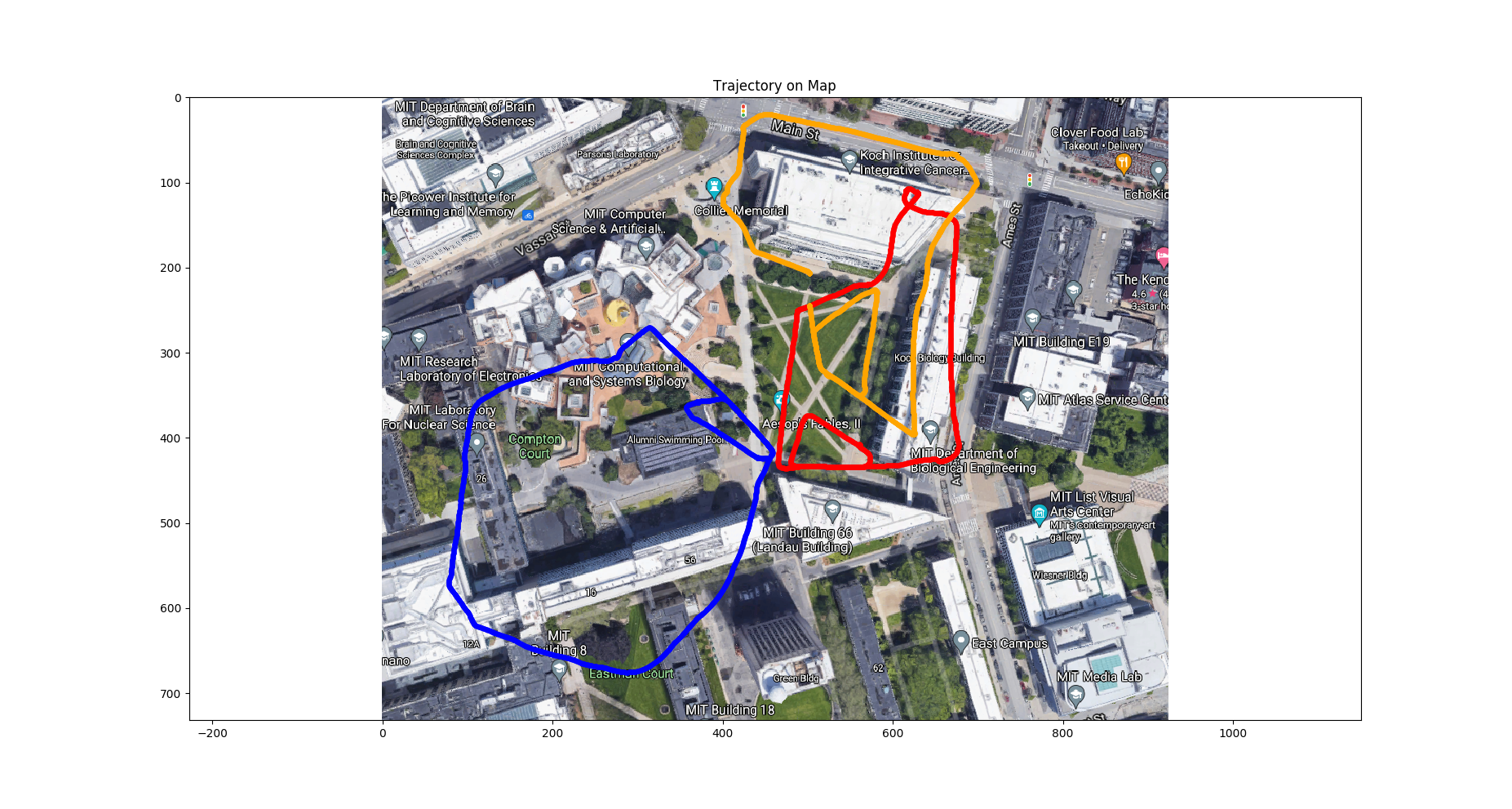}
		\label{fig:stata_multi_traj_approximate}
	}
	\\
	\subfloat[\kimeraMulti (\DGNC with full variable updates)]{%
		\includegraphics[width=0.45\textwidth]{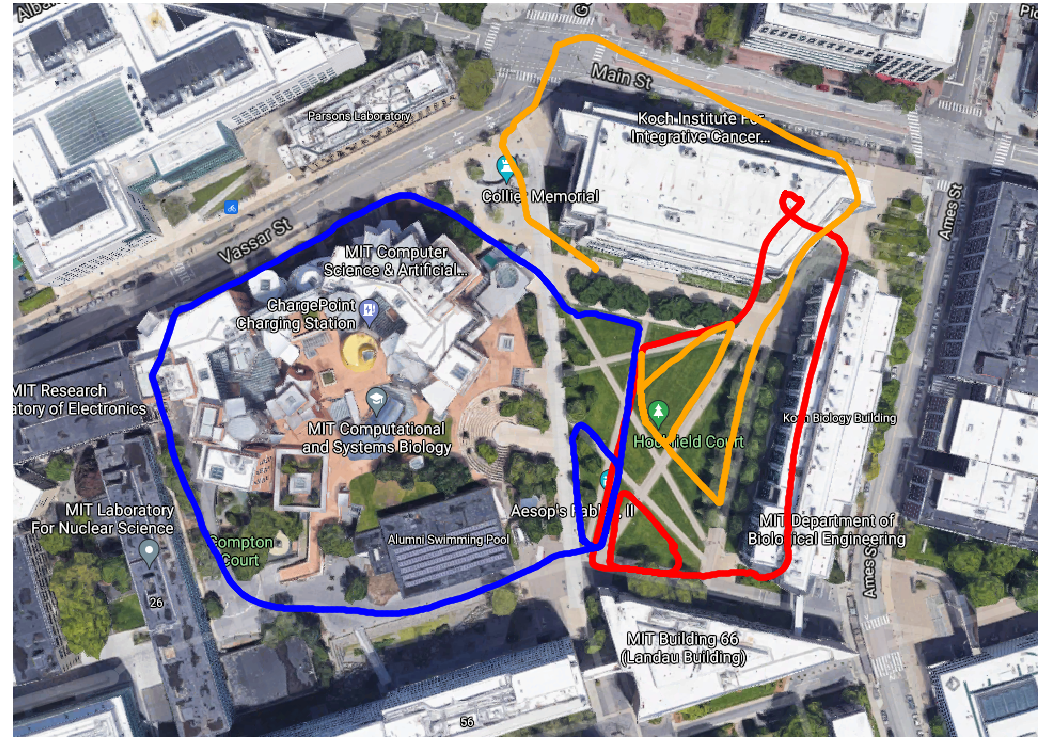}
		\label{fig:stata_multi_traj_full}
	}
	~
	\subfloat[Centralized GNC]{%
		\includegraphics[trim=290 85 290 95, clip, width=0.45\textwidth]{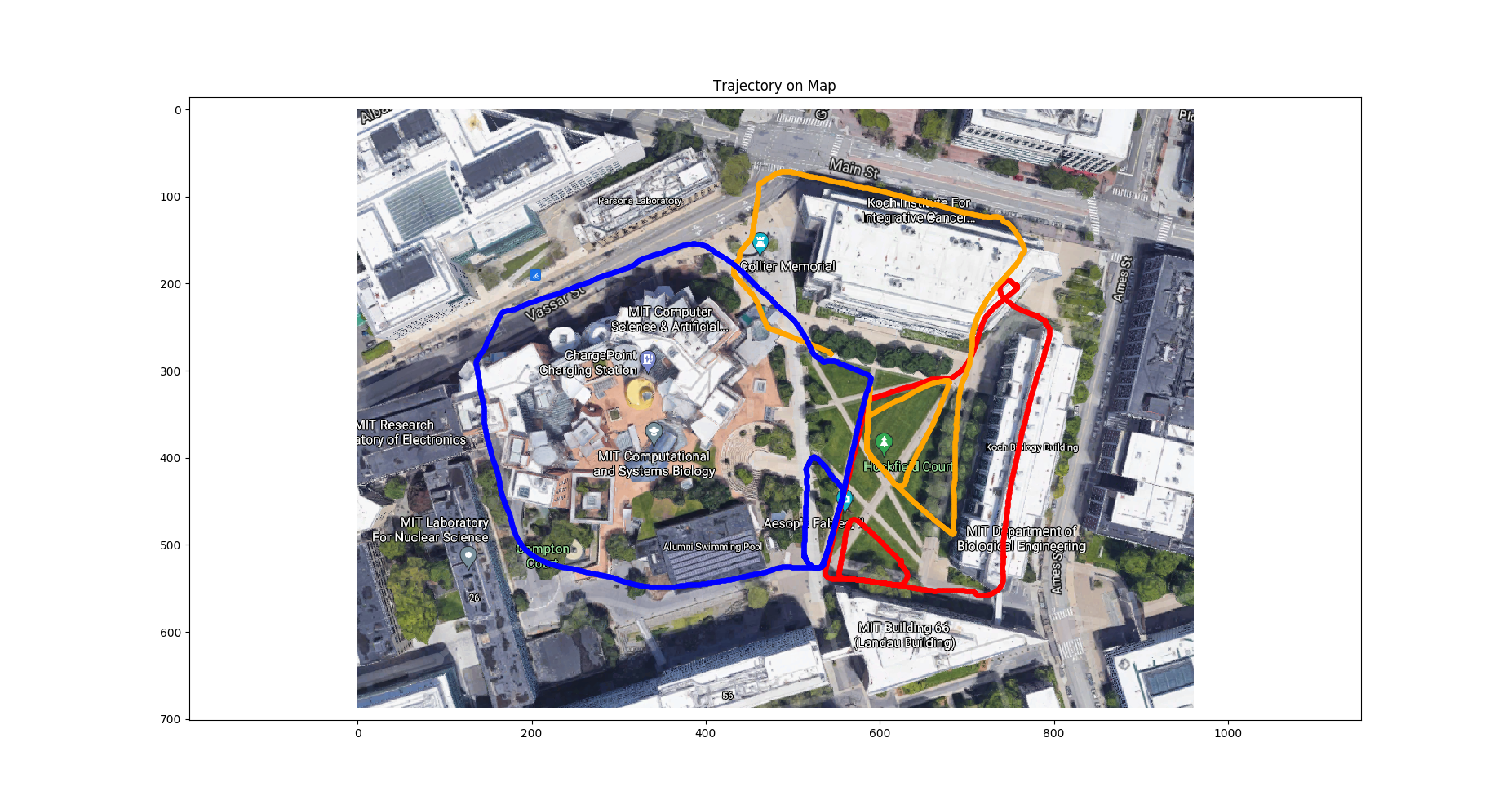}
		\label{fig:stata_multi_traj_centralized}
	}
	\caption{
		\textbf{Stata experiment.}
		(a) Trajectory estimate from \kimeraVIO.
		(b) Trajectory estimate produced by \kimeraMulti, using \DGNC with the default approximate variable updates.
		(c) Trajectory estimate produced by \kimeraMulti, using \DGNC with full variable updates.
		(d) Trajectory estimate produced by centralized GNC.}
	\label{fig:real-world-experiment:stata}
\end{figure*}

\begin{figure}[th]
	\centering
	\includegraphics[trim=25 0 40 40, clip, width=0.46\textwidth]{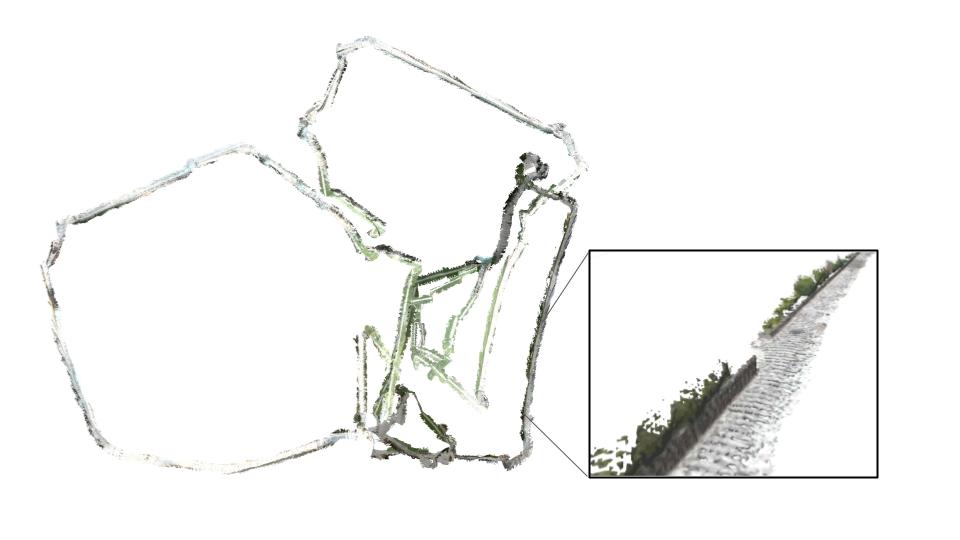}
	\caption{
		\textbf{Stata experiment.}
		Optimized mesh produced by \kimeraMulti corresponding to 
		trajectory estimate shown in Fig.~\ref{fig:stata_multi_traj_full}.
		}
	\label{fig:stata_multi_mesh}
\end{figure}

\begin{table}[t]
	\caption{ \footnotesize
		Loop closure statistics on outdoor datasets.
		For each pair of robots, we show the number of loop closures accepted by \DGNC over the total number of putative loop closures (including outliers).
		Diagonal entries correspond to intra-robot loop closures.
	}
	\label{tab:real-world-loop-closures}
	\centering
	\setlength{\tabcolsep}{1.8pt}
	\renewcommand{\arraystretch}{1.6}
	\subfloat[\medfield experiment]{
		\begin{tabular}{|c|c|c|c|}
			\hline
			& Robot 0                  & Robot 1                  & Robot 2 \\ \hline
			Robot 0 & 1/1                      & 11/41                    & 27/53   \\ \hline
			Robot 1 & \cellcolor[HTML]{C0C0C0} & 79/114                   & 340/707 \\ \hline
			Robot 2 & \cellcolor[HTML]{C0C0C0} & \cellcolor[HTML]{C0C0C0} & 172/182 \\ \hline
		\end{tabular}
		\label{tab:real-world-loop-closures:medfield}}
	~
	\subfloat[\Stata experiment]{
		\begin{tabular}{|c|c|c|c|}
			\hline
			& Robot 0                  & Robot 1                  & Robot 2 \\ \hline
			Robot 0 & 391/416                  & 1/2                      & 1/1 \\ \hline
			Robot 1 & \cellcolor[HTML]{C0C0C0} & 217/271                  & 0/0 \\ \hline
			Robot 2 & \cellcolor[HTML]{C0C0C0} & \cellcolor[HTML]{C0C0C0} & 57/76 \\ \hline
		\end{tabular}
		\label{tab:real-world-loop-closures:stata}}
	\vspace{-20pt}
\end{table}

\begin{table}[t]
	\caption{
		\footnotesize
		Trajectory lengths and
		end-to-end errors in meters on outdoor datasets. \vspace{-2mm} }
	\label{tab:real-world-end-to-end}
	\centering
	\renewcommand{\arraystretch}{1.5}
	\resizebox{.49\textwidth}{!}{
		\begin{tabular}{|c|c|c|c|c|c|}
			\hline
			Dataset   & Robot ID   & Length [m] & \kimeraVIO      & \kimeraMulti     & Centralized  \\ \hline
			& 0          & 600      & 18.74           & 0.01            & 0.01                  \\
			Medfield  & 1          & 860      & 14.84           & 0.13            & 0.13                    \\ 
			& 2          & 728      & 24.55           & 0.09            & 0.09                     \\
			\hline
			& 0          & 515      & 49.02           & 0.03            & 0.01                    \\
			Stata     & 1          & 570      & 24.19           & 33.13           & 21.56                    \\ 
			& 2          & 610      & 29.35           & 1.26            & 1.17                     \\
			\hline
	\end{tabular}}
	\vspace{-3mm}
\end{table}

On the \medfield dataset (Fig.~\ref{fig:real-world-experiment:medfield}), \kimeraVIO accumulates a drift of approximately 15-25~m on each trajectory sequence.
We note that the drift is mostly in the vertical direction, hence only partially visible in Fig.~\ref{fig:medfield_vio_traj}. 
Through loop closures and robust distributed PGO, 
\kimeraMulti significantly reduces the error and furthermore achieves the same performance as the centralized solver as shown in Table~\ref{tab:real-world-end-to-end}.
In this case, the global pose graph has 15650 poses in total (including all robots). \DGNC uses a total of 100 \RBCD iterations which takes 53 seconds.
Further runtime reduction may be achieved by decreasing the rate at which keyframes are created. 
In summary, our trajectory estimation results together with the final optimized mesh shown in Fig.~\ref{fig:medfield_multi_mesh} demonstrate the effectiveness of the proposed system.

In comparison, the \Stata dataset (Fig.~\ref{fig:real-world-experiment:stata}) is more challenging, partially due to the lack of enough inter-robot loop closures (Table~\ref{tab:real-world-loop-closures:stata}).
\kimeraVIO accumulates higher drifts as shown in Fig.~\ref{fig:stata_vio_traj} and Table~\ref{tab:real-world-end-to-end}.
Fig.~\ref{fig:stata_multi_traj_approximate} shows the \kimeraMulti trajectory estimates produced using the default settings of \DGNC (see Algorithm~\ref{alg:dgnc}).
In this case, the global pose graph has 11184 poses in total.
\DGNC uses 120 \RBCD iterations which takes 50 seconds.
We observe that while the orange and red trajectory estimates are qualitatively correct, the blue trajectory is not correctly aligned in the global frame.
This is because with the approximate variable updates of \DGNC (presented in Section~\ref{sec:dpgo}), the only inter-robot loop closure with the blue trajectory is rejected. 
Additionally, with fewer inter-robot loop closures, \RBCD generally converges at a slower rate.
To resolve this issue, we increase the number of \RBCD iterations within each variable update, hence making \DGNC more similar to the centralized GNC algorithm.
With this change, \DGNC uses a total of 2000 \RBCD iterations which takes 14 minutes.
However, the final trajectory estimates (Fig.~\ref{fig:stata_multi_traj_full}) are significantly improved and are close to centralized GNC (Fig.~\ref{fig:stata_multi_traj_centralized}).
The corresponding end-to-end errors are also close to centralized GNC as shown in Table~\ref{tab:real-world-end-to-end}.
Fig.~\ref{fig:stata_multi_mesh} shows the optimized mesh produced by \kimeraMulti corresponding to the trajectory estimates shown in Fig.~\ref{fig:stata_multi_traj_full}.
In this experiment, the difficulty faced by \kimeraMulti is primarily due to the lack of inter-robot loop closures (Table~\ref{tab:real-world-loop-closures:stata}).
In the future, we plan to further improve the loop closure detection module to gain better performance in similar visually challenging scenarios.


\section{Conclusion}
We presented \kimeraMulti, \emph{a distributed multi-robot system for robust and dense metric-semantic SLAM.}
Our system advances state-of-the-art multi-robot perception 
by estimating 3D mesh models that capture both dense geometry and semantic information of the environment.
\kimeraMulti is \emph{fully distributed}:
each robot performs independent navigation, using \kimera to estimate local trajectories and meshes in real-time.
When communication becomes available, 
robots engage in \emph{local communication} to detect loop closures and perform distributed trajectory estimation. 
From the globally optimized trajectory estimates, each robot performs local mesh optimization to refine its local map. 
We also presented \DGNC, a novel two-stage method for \emph{robust} distributed pose graph optimization, which serves as the estimation backbone of \kimeraMulti and outperforms prior outlier rejection methods.

We performed extensive evaluation of \kimeraMulti, using a combination of photo-realistic simulations, indoor SLAM benchmarking datasets, and large-scale outdoor datasets.
Our results demonstrated that \kimeraMulti
(i) provides robust and accurate trajectory estimation while being fully distributed,
(ii) estimates 3D meshes with improved metric-semantic accuracy compared to inputs from \kimera,
and (iii) is communication-efficient and achieves significant communication reductions compared to baseline centralized systems.

\section*{Acknowledgments}
The authors gratefully acknowledge Dominic Maggio for assistance during outdoor data collection.

\bibliographystyle{IEEEtran}
\bibliography{./includes/refs.bib,./includes/myRefs.bib,./includes/cslam.bib}

\vspace{-10mm}
\begin{IEEEbiography}[{\includegraphics[width=1in,height=1.25in,clip,keepaspectratio]
		{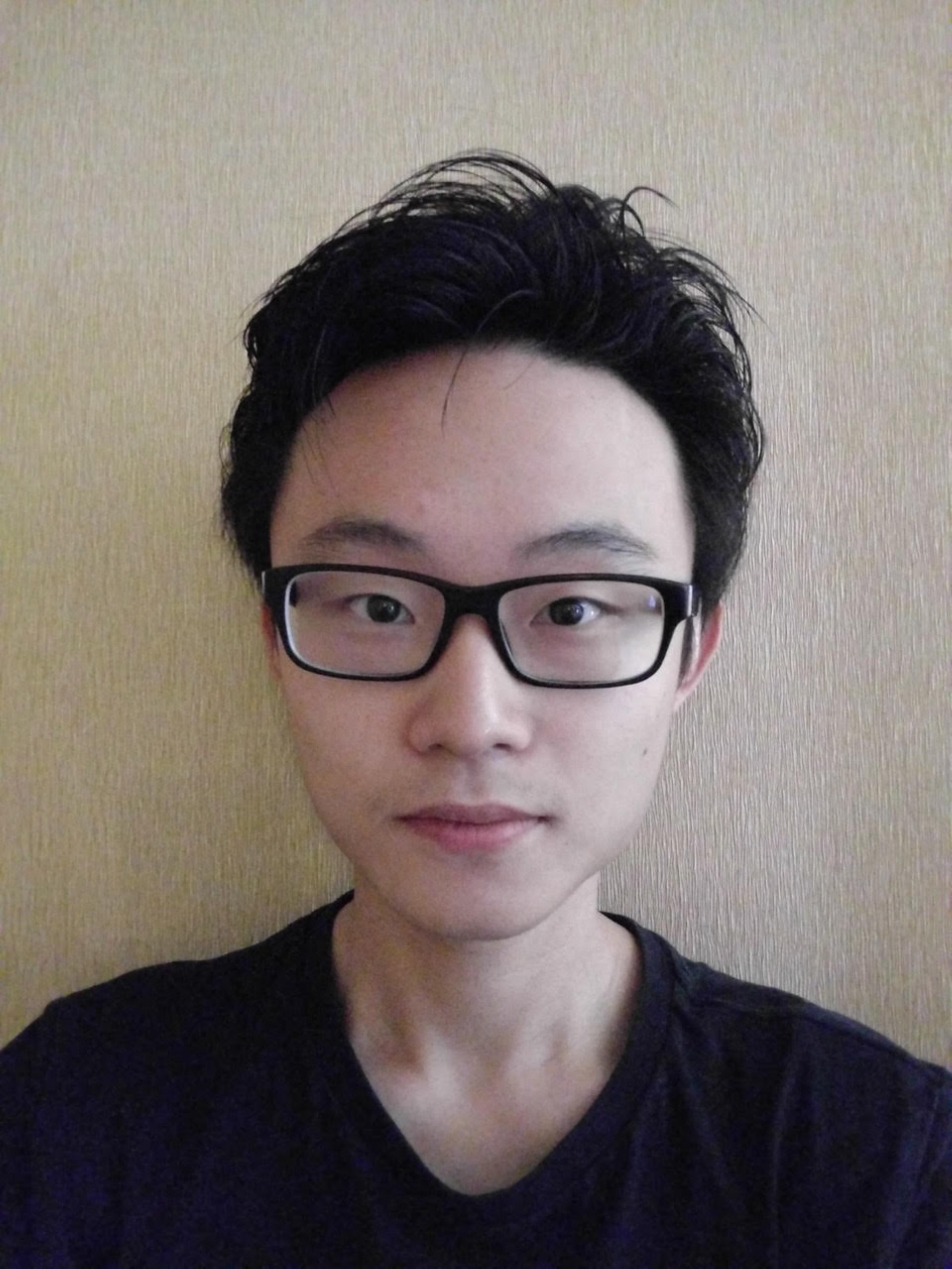}}]{Yulun Tian}
	received the B.A. degree in computer science from UC Berkeley, Berkeley, CA, USA, in 2017, and the S.M. degree in aeronautics and astronautics in 2019 from the Massachusetts Institute of Technology, Cambridge, MA, USA, where he is currently working toward the Ph.D. degree in aeronautics and astronautics. 
	His work received a 2020 Honorable Mention from the IEEE Robotics and Automation Letters.
	
	His current research interest includes distributed optimization and estimation with applications to localization and mapping in multi-agent systems.
\end{IEEEbiography}
\vspace{-10mm}
\begin{IEEEbiography}[{\includegraphics[width=1in,height=1.25in,clip,keepaspectratio]
		{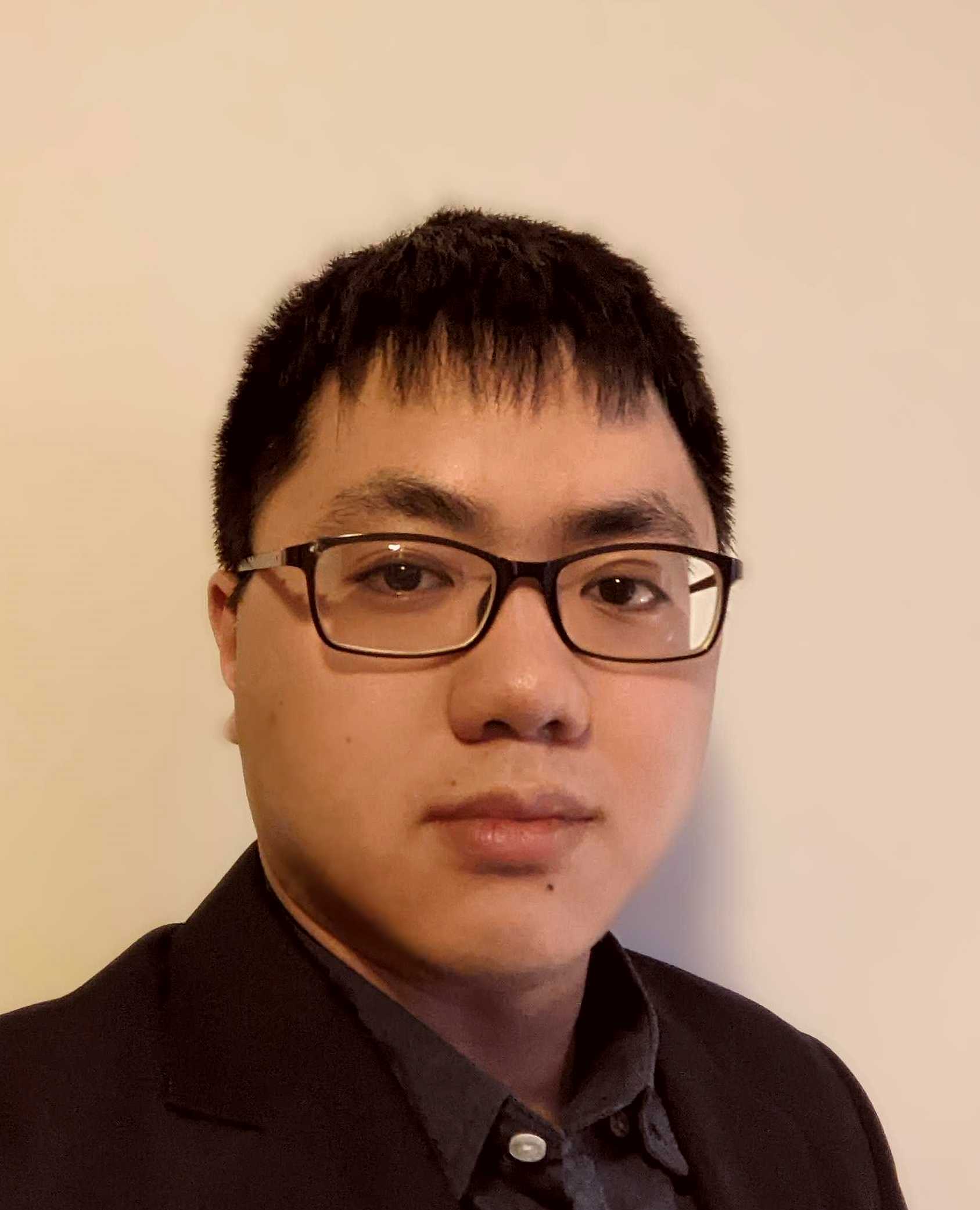}}]{Yun Chang}
	is a PhD student in the Department of Aeronautics and Astronautics and the Laboratory for
	Information \& Decision Systems (LIDS). He is currently a member of SPARK lab, led by Professor Luca Carlone. He has obtained a B.S. degree in Aerospace Engineering from the Massachusetts Institute of Technology in 2019 and a M.S. in Aeronautics and Astronautics from the Massachusetts Institute of Technology in 2021. His research interest includes robust localization and mapping with applications to multirobot systems. He is a recipient of the MIT AeroAstro Andrew G. Morsa Memorial Award (2019) and the Henry Webb Salisbury Award (2019).
\end{IEEEbiography}
\vspace{-10mm}
\begin{IEEEbiography}[{\includegraphics[trim=120 0 320 0, width=1in,height=1.25in,clip,keepaspectratio]
		{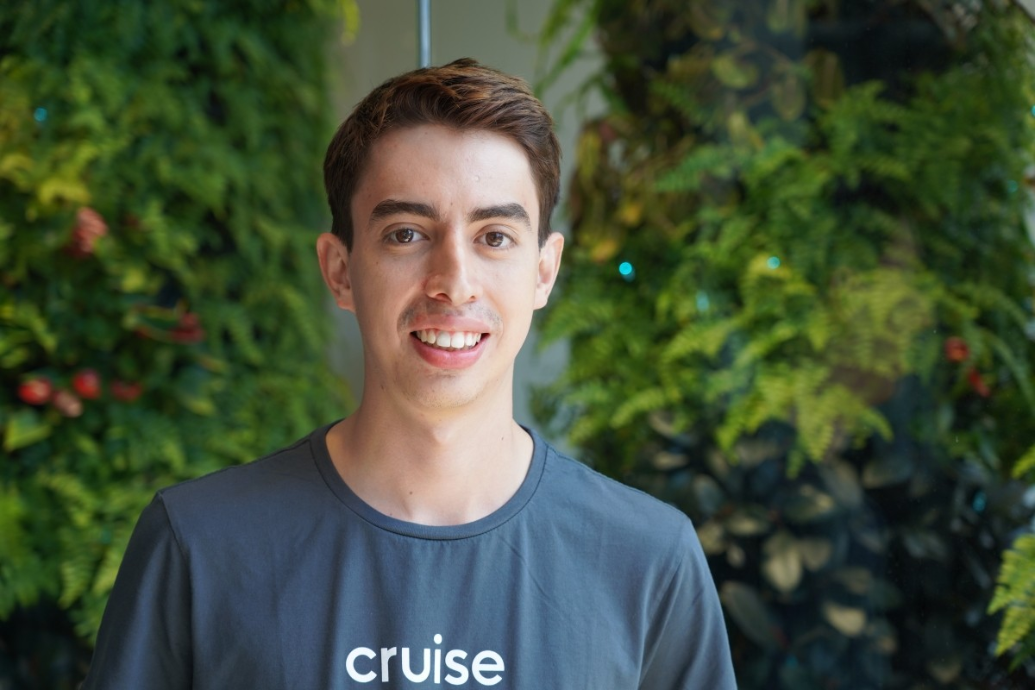}}]{Fernando Herrera Arias}
	is a software engineer at Cruise. He graduated from the Massachusetts Institute of Technology with a S.B. degree in Computer Science and Engineering in 2020 and a M.Eng. in Electrical Engineering and Computer Science in 2021. He is a former member of SPARK lab, led by Professor Luca Carlone. As part of this team, his work included the evaluation of neural networks for loop-closure detection in SLAM systems. His current work is on vehicle dynamics for self-driving cars.
\end{IEEEbiography}

\begin{IEEEbiography}[{\includegraphics[trim=15 0 0 0,width=1in,height=1.25in,clip,keepaspectratio]
		{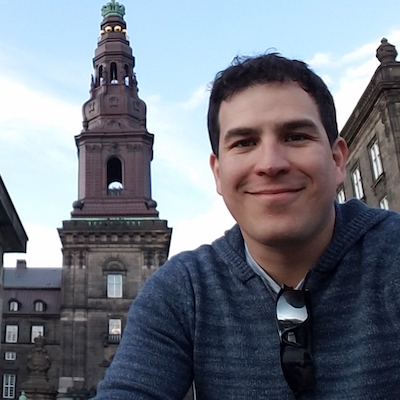}}]{Carlos Nieto-Granda}
	is a Postdoctoral-Fellow in the Computational and Information Sciences Directorate at the U.S. Army Research Laboratory (DEVCOM/ARL). He has obtained a B.S. degree in Electronics Systems from Tecnol{\'o}gico de Monterrey, Campus Estado de Mexico, Mexico; an M.S. degree in Computer Science from Georgia Institute of Technology; and a Ph.D. degree in Intelligent Systems, Robotics, and Control from University of California San Diego. His research interests include autonomous exploration, coordination, and decision-making for heterogeneous multi-robot teams focused on state estimation, sensor fusion, computer vision, localization and mapping, autonomous navigation, and control in complex environments.
\end{IEEEbiography}

\begin{IEEEbiography}[{\includegraphics[trim=300 0 300 0, width=1in,height=1.25in,clip,keepaspectratio]
		{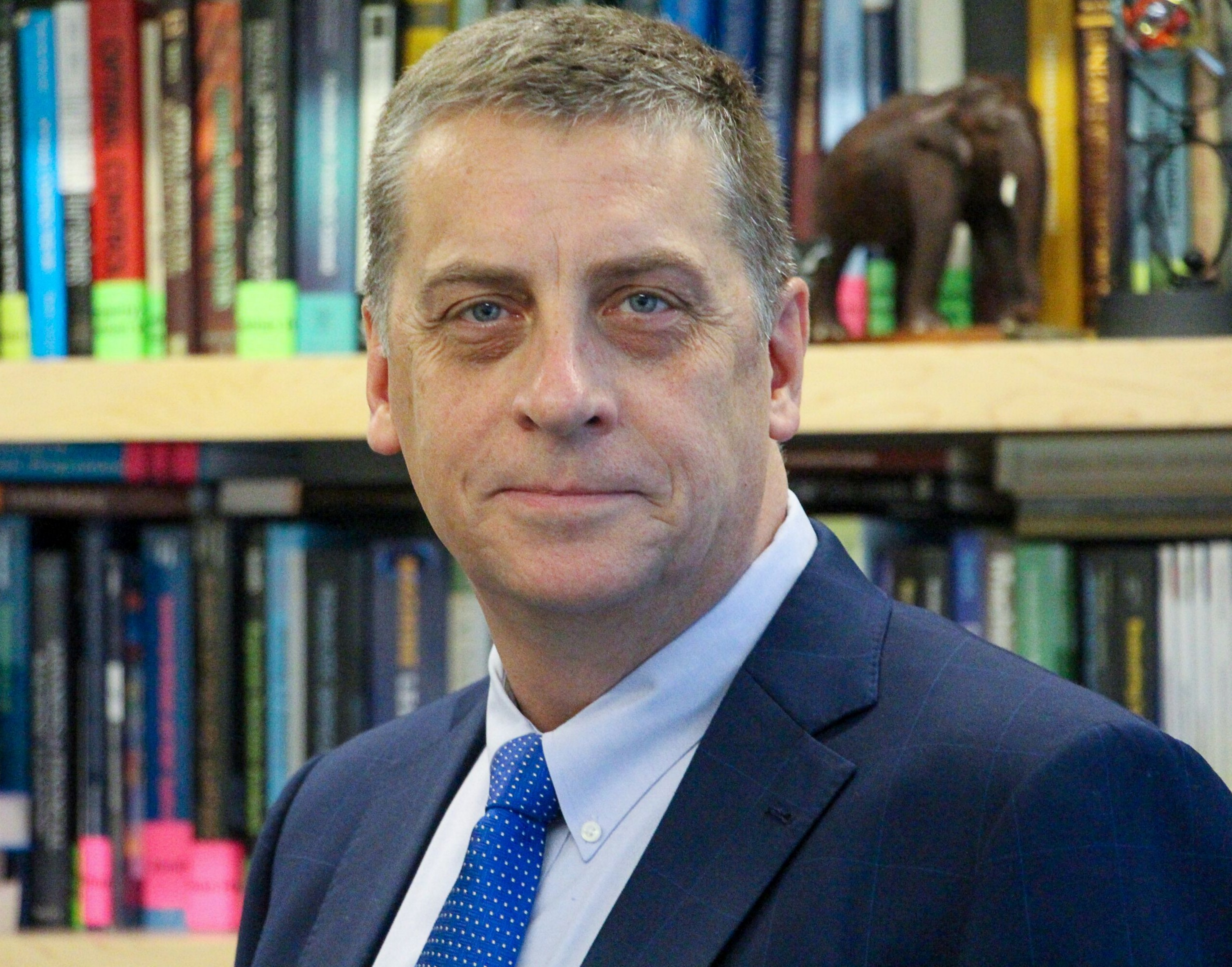}}]{Jonathan P. How}
	(Fellow, IEEE) received the
	B.A.Sc. degree from the University of Toronto (1987), and the S.M. and Ph.D. degrees in aeronautics and astronautics from MIT (1990 and 1993). Prior to joining MIT in 2000,
	he was an Assistant Professor at Stanford University. He is currently the
	Richard C. Maclaurin Professor of aeronautics and astronautics at MIT. Some of his awards include the IEEE CSS Distinguished Member Award (2020), AIAA Intelligent Systems Award (2020), 
	IROS Best Paper Award on Cognitive Robotics (2019), and the AIAA Best
	Paper in Conference Awards (2011, 2012, 2013). 
	He was the Editor-in-chief of IEEE Control Systems Magazine (2015--2019), is a Fellow of AIAA, and 
	was elected to the National Academy of Engineering in 2021.
\end{IEEEbiography}

\begin{IEEEbiography}[{\includegraphics[width=1in,height=1.25in,clip,keepaspectratio]
		{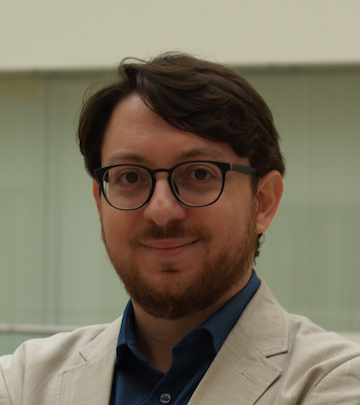}}]{Luca Carlone}
	is the Leonardo Career Development Associate Professor in the Department of Aeronautics and Astronautics at the Massachusetts Institute of Technology, and a Principal Investigator in the Laboratory for Information \& Decision Systems (LIDS). He has obtained a B.S. degree in mechatronics from the Polytechnic University of Turin, Italy, in 2006; an S.M. degree in mechatronics from the Polytechnic University of Turin, Italy, in 2008; an S.M. degree in automation engineering from the Polytechnic University of Milan, Italy, in 2008; and a Ph.D. degree in robotics also the Polytechnic University of Turin in 2012. He joined LIDS as a postdoctoral associate (2015) and later as a Research Scientist (2016), after spending two years as a postdoctoral fellow at the Georgia Institute of Technology (2013-2015). His research interests include nonlinear estimation, numerical and distributed optimization, and probabilistic inference, applied to sensing, perception, and decision-making in single and multi-robot systems. His work includes seminal results on certifiably correct algorithms for localization and mapping, as well as approaches for visual-inertial navigation and distributed mapping. He is a recipient of the Best Student Paper Award at IROS 2021, the Best Paper Award in Robot Vision at ICRA 2020, a 2020 Honorable Mention from the IEEE Robotics and Automation Letters, a Track Best Paper award at the 2021 IEEE Aerospace Conference, the 2017 Transactions on Robotics King-Sun Fu Memorial Best Paper Award, the Best Paper Award at WAFR 2016, the Best Student Paper Award at the 2018 Symposium on VLSI Circuits, and he was best paper finalist at RSS 2015 and RSS 2021. He is also a recipient of the NSF CAREER Award (2021), the RSS Early Career Award (2020), the Google Daydream (2019) and the Amazon Research Award (2020), and the MIT AeroAstro Vickie Kerrebrock Faculty Award (2020).
\end{IEEEbiography}

\end{document}